\documentclass{article}

\PassOptionsToPackage{numbers, compress}{natbib}
\usepackage[preprint]{neurips_2026}

\usepackage[utf8]{inputenc}
\usepackage[T1]{fontenc}
\usepackage{hyperref}
\hypersetup{hidelinks}
\usepackage{url}
\usepackage{booktabs}
\usepackage{amsfonts}
\usepackage{amsmath}
\usepackage{amssymb}
\usepackage{nicefrac}
\usepackage{microtype}
\usepackage[table]{xcolor}
\usepackage{graphicx}
\usepackage{wrapfig}
\usepackage{multirow}
\usepackage{enumitem}
\usepackage{array}
\usepackage{tabularx}
\usepackage{capt-of}
\usepackage{needspace}

\title{MUSE: Agentic 3D Scene Authoring via Memory-Grounded Incremental Requirement Satisfaction}

\author{%
  Ruijie Xu\textsuperscript{1,*}
  \And
  Xinnan Zhu\textsuperscript{1,*}
  \And
  Jiayu Ying\textsuperscript{1}
  \And
  Daoguo Dong\textsuperscript{2}
  \And
  Yuzhou Ji\textsuperscript{3}
  \And
  Xin Tan\textsuperscript{1,\textdagger}
  \AND
  \normalfont
  \textsuperscript{1}East China Normal University \\
  \textsuperscript{2}Fudan University \\
  \textsuperscript{3}Shanghai Jiao Tong University \\
  \textsuperscript{*}Equal contribution. \quad
  \textsuperscript{\textdagger}Corresponding author.
}

\newcounter{algorithm}

\newcommand{\relatedworkpara}[1]{\par\smallskip\noindent\textbf{#1.}\enspace\ignorespaces}
\newcolumntype{Y}{>{\raggedright\arraybackslash}X}

\begin{document}

\maketitle

\begin{center}
  \textbf{Project page:} \url{https://xujay111.github.io/muse_githubio/}
\end{center}

\begin{abstract}
Text-driven 3D scene generation is a promising technique for digital content creation, embodied AI simulation, and interactive design, yet practical workflows often require refining, extending, or correcting existing scenes while preserving non-target content. Existing methods can produce realistic and structurally plausible scenes, but they generally lack editability with requirement-level state tracking, so part-level failures often lead to full-scene regeneration or manual intervention. To tackle this challenge, we formulate controllable 3D scene authoring as incremental requirement satisfaction, unifying construction and editing. In this paper, we present \textsc{MUSE}, a memory-grounded multi-agent framework in which an Architect compiles instructions into structured requirements, a Sculptor executes local scene operations, and an Inspector verifies each step while updating Working, Scene, and Skill Memory. 
To evaluate requirement-level controllability and preservation-aware editing, we introduce AuthorBench, offering 145 constrained construction cases and a 1,584-case preservation-aware editing pool paired with external structured checks. On full construction cases, \textsc{MUSE} improves All-Goal success from 37.9 to 80.7 and surface-constraint fulfillment from 35.0 to 92.6 over the strongest baseline. On a stratified 240-case editing test split, \textsc{MUSE} achieves 49.6 All-Goal success, 99.9 preservation rate, and only 0.6 unintended change rate. Beyond automated metrics, human evaluations on compared local-editing baselines support stronger alignment with user intent, and downstream navigation-proxy tests indicate stronger spatial stability. Combined with ablations validating our memory designs, these results establish MUSE as an effective framework for controllable 3D scene authoring.
\end{abstract}

\section{Introduction}
\label{sec:introduction}
\begin{figure*}[t]
    \centering
    \includegraphics[width=\textwidth]{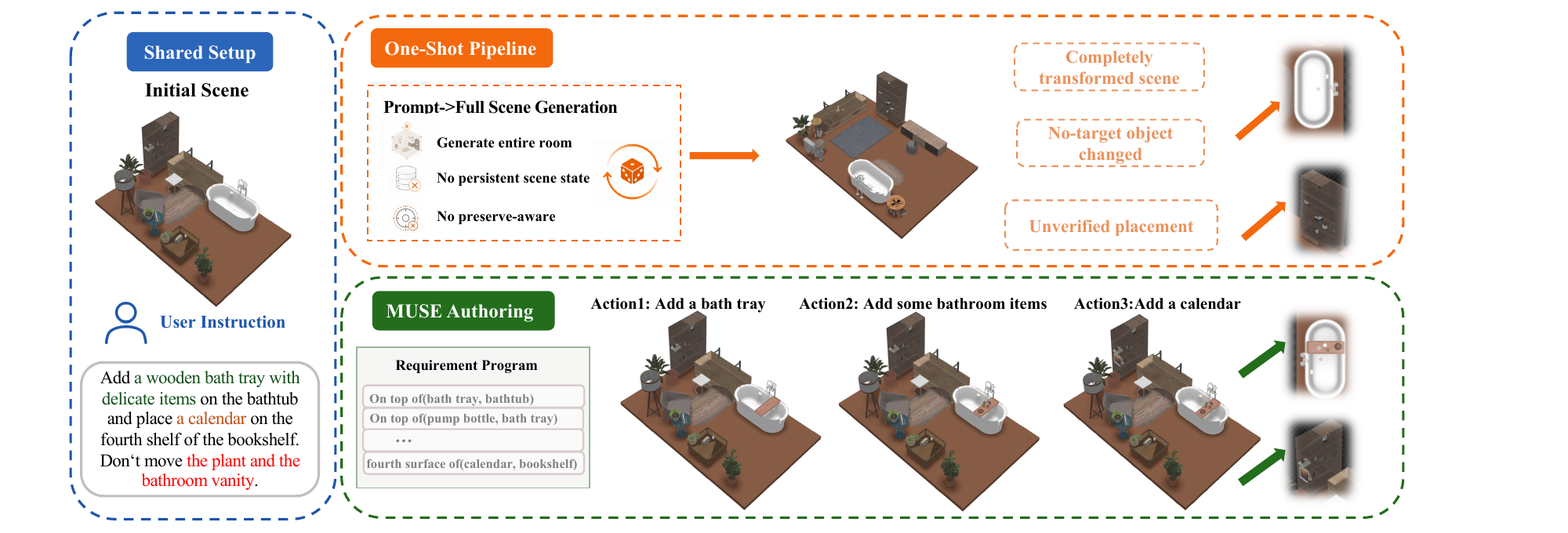}
    \vspace{0.3em}
    \caption{
    \textbf{From one-shot regeneration to persistent 3D scene authoring.}
    One-shot pipelines regenerate the whole scene and may disturb non-target regions, whereas \textsc{MUSE} performs verified local actions over persistent memory to preserve satisfied content.
    }
    \label{fig:teaser}
\end{figure*}

\vspace{0.3em}

Text-driven 3D scene generation enables users to create complex three-dimensional environments from natural language, with broad applications in simulation, game design ~\cite{yang2024holodeck,yang2024physcene,fu2024anyhome,gu2025artiscene,zhang2026sage}. However, many practical settings, such as embodied AI training and interactive design, require more than one-shot synthesis: users often need to add, remove, or adjust local content in an existing scene while keeping non-target regions structurally and semantically stable. Supporting this workflow requires controllable generation, where a system maintains an explicit, persistent representation of the scene state and generation intent to guide subsequent local operations.

Despite substantial progress in spatial modeling, existing text-driven 3D scene generation methods still fall short of controllable generation. They can be grouped into three families. Data-driven methods~\cite{fu20213dfront,paschalidou2021atiss,wang2021sceneformer,dhamo2021graphto3d,tang2024scenegeneration,lin2024instructscene} rely on implicit generation processes, making it difficult to explicitly control local structures. LLM-based planning methods~\cite{feng2024layoutgpt,yang2024holodeck,sun2025layoutvlm} improve open-vocabulary flexibility, but rely on one-shot context without persistently tracking execution history or unmet constraints. Agentic methods introduce tool use and closed-loop reflection~\cite{yang2025sceneweaver,wu2026scenesmith,zhang2026sage}, yet most of them still optimize within a single generation round rather than manage scene state across interactions. Consequently, existing systems generally lack explicit modeling and binding of requirement-level states, making it difficult to repair local failures without disturbing non-target content. In practice, when a local result is unsatisfactory, they often resort to global regeneration instead of truly controllable local revision.

Therefore, scene editability is central to controllable 3D generation. Practical workflows often do not start from an empty scene, but instead refine, extend, or correct an existing one. Accordingly, rather than regenerating globally after local failures, an ideal system should reuse semantic relations, satisfied bindings, and intermediate scene state, while applying a shared set of atomic authoring actions to update only the target region and preserve non-target content. Based on this observation, we reformulate 3D scene synthesis as a controllable authoring process that unifies generation and editing. Centered on persistent state, it supports both initial construction and fine-grained editing through requirement decomposition, local execution, verification, and memory mechanisms.

Realizing this incremental authoring process requires structured memory. \emph{Working Memory} tracks requirement status, diagnostic signals, and protected bindings, preventing subsequent operations from breaking satisfied content. \emph{Scene Memory} maintains a persistent hierarchical representation of the 3D environment to enable stepwise spatial reasoning. \emph{Skill Memory} stores reusable decomposition patterns for recurring scene motifs, reducing repeated decomposition errors.

Based on this design, we present \textsc{MUSE}, a multi-agent framework for unified 3D scene authoring. For each instruction, the \emph{Architect} compiles natural language into structured requirement programs, the \emph{Sculptor} incrementally executes them, and the \emph{Inspector} verifies each step and updates memory. This tight integration ensures \textsc{MUSE} reliably identifies unmet requirements, protects satisfied content, and maintains a persistent scene state across continuous interactions.

To evaluate this perspective, we introduce AuthorBench, a comprehensive benchmark for requirement-level 3D scene authoring. It features 145 constrained construction cases and a 1,584-case preservation-aware editing set, paired with structured checks to assess fine-grained spatial and semantic constraints. Experiments in construction show \textsc{MUSE} substantially outperforms baselines, improving All-Goal construction success from 37.9 to \textbf{80.7} and surface fulfillment from 35.0 to \textbf{92.6}. In editing, it achieves \textbf{49.6} All-Goal success with near-perfect locality (\textbf{99.9} preservation, \textbf{0.6} unintended changes). Moreover, human evaluations and downstream navigation-proxy tests support its alignment and spatial stability. Finally, ablations validate the role of verification and memory mechanisms in controllable authoring. In summary, our contributions are: 

\begin{itemize}[nosep,leftmargin=*]
    \item We formulate controllable 3D scene authoring as incremental requirement satisfaction over persistent scene state, establishing a unified requirement-level foundation for construction and editing through decomposition, execution, verification, and preservation.
    \item We present \textsc{MUSE}, a memory-grounded multi-agent framework in which Architect, Sculptor, and Inspector agents operate over Working, Scene, and Skill Memory to support local revision, protection of satisfied content, and reusable decomposition of recurring scene motifs.
    \item We introduce AuthorBench, a unified requirement-level benchmark for evaluating both scene construction and preservation-aware editing. Extensive evaluations across automated metrics, human preference studies, downstream navigation tests, and ablations demonstrate that \textsc{MUSE} achieves strong performance against recent baselines in controllable 3D scene authoring.
\end{itemize}

\section{Related work}
\label{sec:related_work}

\relatedworkpara{Text-driven 3D scene synthesis} Prior work on text-driven 3D scene synthesis spans three main lines. Data-driven methods capture object distributions and relations from indoor scene datasets~\cite{fu20213dfront,paschalidou2021atiss,wang2021sceneformer,dhamo2021graphto3d,tang2024scenegeneration,lin2024instructscene}, but offer limited explicit control over local structure and intermediate decisions. LLM-based planning methods improve open-vocabulary scene construction through spatial programs or layout optimization~\cite{feng2024layoutgpt,yang2024holodeck,sun2025layoutvlm}, while agentic frameworks introduce tool use and iterative refinement~\cite{yang2025sceneweaver,wu2026scenesmith,zhang2026sage}. However, these methods are fundamentally restricted to one-shot generation from a prompt, making it difficult to isolate failed constraints, revise only the affected regions, and preserve previous results during continuous authoring. To address this issue, we introduce controllable 3D scene authoring, which unifies construction and editing at the requirement level.

\relatedworkpara{Language-guided 3D scene editing} Existing language-guided 3D scene editing methods can be divided into two broad categories. Conversational or LLM-driven systems support natural-language object addition, removal, and feedback-based adjustment~\cite{yang2024llplace,wang2025chat2layout}, but often rely on implicit reasoning for constraint satisfaction and object protection. Structured methods improve output organization through scene-token, scene-graph, or action-sequence prediction~\cite{bucher2025respace,cheng2026layoutr1}, yet still tend to handle complex edits as holistic commands rather than as independently verifiable requirements. In our setting, the goal is to satisfy a new instruction in a preservation-aware manner, where non-target content should remain stable. Recent planning-based editors move in this direction~\cite{huang2026editasact}, but still do not provide a unified mechanism for tracking progress and preservation during local revision. To overcome these limitations, we formulate editing as memory-grounded incremental requirement satisfaction, leveraging structured memory to explicitly track progress and preserve non-target content.

\relatedworkpara{Memory and structured state in LLM agents} Explicit memory and structured state have proven useful for long-horizon reasoning and multi-step decision making in embodied navigation, LLM agents, and scene-graph reasoning, including spatial maps, open-vocabulary scene graphs, skill libraries, memory streams, and intermediate reasoning traces~\cite{chaplot2020objgoal,gu2023conceptgraphs,wang2023voyager,park2023generativeagents,shinn2023reflexion,yao2023react,huang2022innermonologue,sumers2023coala}. In 3D scene synthesis, recent methods also support iterative feedback or multi-turn interaction during scene synthesis~\cite{yang2025sceneweaver,wu2026scenesmith,zhang2026sage,zhao2026scenerevis}. However, systems such as SceneReVis~\cite{zhao2026scenerevis} typically still rely on dialogue history or model context as their core state. This implicitly restricts their ability to track satisfied constraints, isolate preservable content, and execute safe local modifications. Our work therefore introduces task-structured memory for controllable 3D scene authoring, keeping requirement progress, preservable content, and editable scene structure explicit across interactions.

\section{Method}
\label{sec:method}

\begin{figure}[t]
  \centering
  \includegraphics[width=1\linewidth]{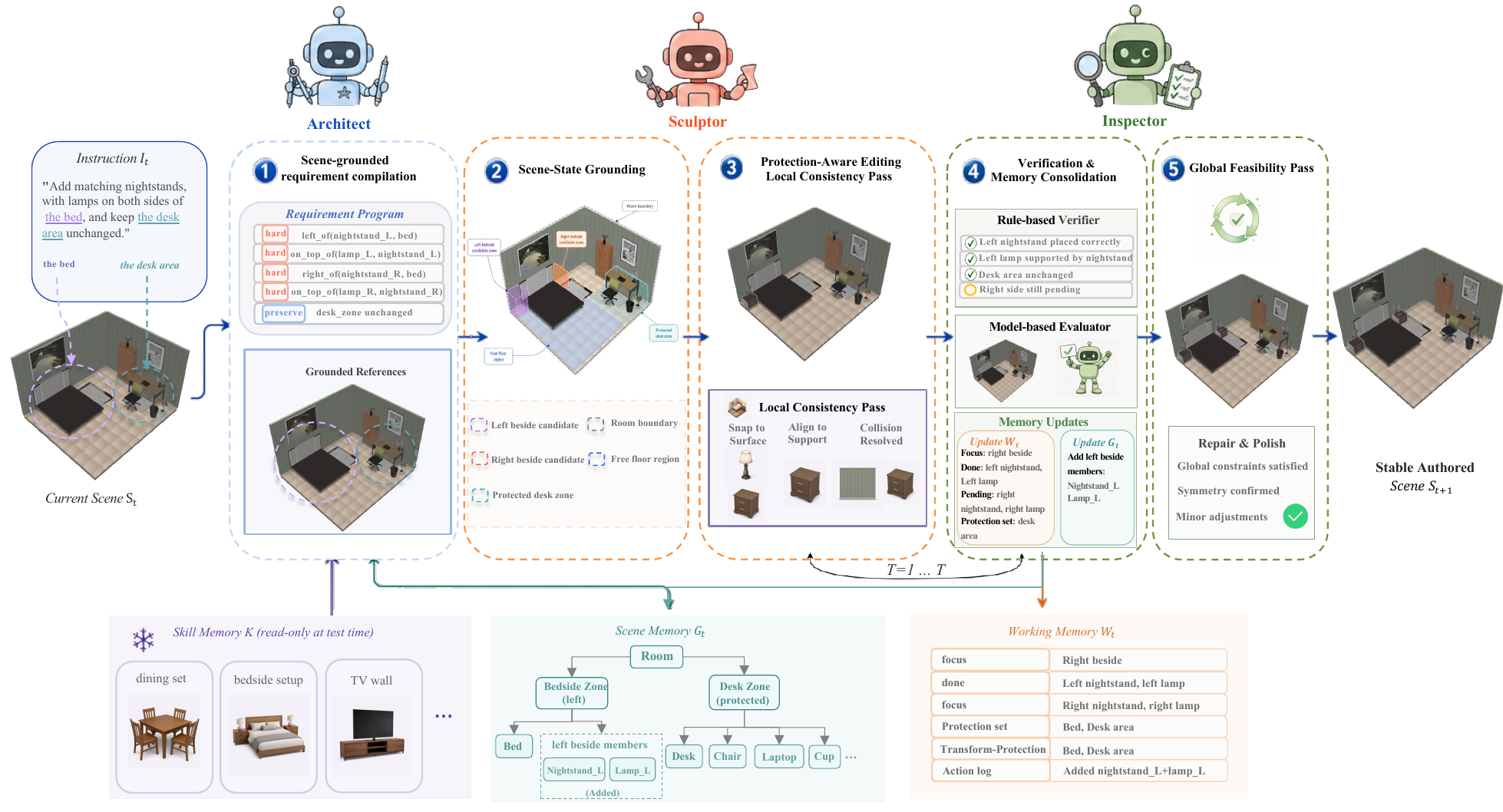}
  \vspace{0.3em}
  \caption{\textbf{Overview of \textsc{MUSE}.} For each interaction $t$, \textsc{MUSE} follows a five-stage authoring loop: \textbf{(1)} the Architect compiles $I_t$ with Scene Memory $\mathcal{G}_t$ and Skill Memory $\mathcal{K}$ into a requirement program $\mathcal{B}_t$; \textbf{(2)} the Sculptor grounds $\mathcal{S}_t$ into an editable room state; \textbf{(3)} the Sculptor executes focused requirements with protection-aware local repair; \textbf{(4)} the Inspector verifies requirement satisfaction and consolidates Working and Scene Memory; and \textbf{(5)} a global feasibility pass resolves residual conflicts and returns $\mathcal{S}_{t+1}$ with updated persistent memory.}
  \label{fig:overview}
\end{figure}

We present \textsc{MUSE}, a multi-agent framework that realizes controllable 3D scene authoring through persistent requirement-level state. Given either a complex construction prompt or a targeted edit instruction, \textsc{MUSE} compiles the input into a structured requirement program, executes requirements locally, verifies each step, and updates memory to preserve satisfied content. Figure~\ref{fig:overview} presents the full five-step pipeline. Below, we first formulate scene authoring as incremental requirement satisfaction, then describe the Architect, Sculptor, and Inspector modules.

\subsection{Task formulation}
\label{sec:task}

We formulate \emph{controllable 3D scene authoring} as incremental requirement satisfaction over a persistent scene state. At interaction $t$, the system receives a current scene $\mathcal{S}_t$, a natural-language input $I_t$ (either a construction prompt or a targeted edit instruction), and the memory state $\mathcal{M}_t$. This input is first compiled into a structured requirement program $\mathcal{B}_t=\{r_i\}_{i=1}^{n}$. Driven by $\mathcal{B}_t$, the system iteratively produces a sequence of atomic operations
\begin{equation}
\label{eq:task}
    \Delta_t = (\delta_1,\ldots,\delta_m), \qquad
    \mathcal{S}_{t+1} = \mathcal{S}_t \oplus \Delta_t,
\end{equation}
where $\oplus$ signifies the sequential execution of these operations on the scene state. Each operation $\delta_j \in \Delta_t$ instantiates an atomic authoring action, such as object insertion, removal, transformation, or support-aware placement. The objective is \textbf{incremental requirement satisfaction}: instead of resolving the dense program simultaneously, the system iteratively schedules and fulfills requirements until every $r_i \in \mathcal{B}_t$ is verified.

To keep this incremental loop stable, the system relies on a three-layer memory state $\mathcal{M}_t = (\mathcal{W}_t,\; \mathcal{G}_t,\; \mathcal{K})$. \emph{Working Memory ($\mathcal{W}_t$)} is instantiated per-input, storing the execution focus, satisfaction status, role bindings, and protection sets to track progress and prevent regressions. \emph{Scene Memory ($\mathcal{G}_t$)} persists across interactions. By maintaining the room shell, support surfaces, functional zones, and the core object graph, it grounds spatial references, supports local repairs, and ensures the scene structure remains editable. Finally, \emph{Skill Memory ($\mathcal{K}$)} acts as a reusable library of motif priors, decomposition cues, and failure warnings to improve the parsing of recurring scene configurations.

Under this formulation, construction and editing are unified as two operating modes. Construction starts from an empty or partially specified room shell and incrementally realizes requirements to build a new scene. Editing starts from an existing scene and incorporates preservation constraints into the requirement program, ensuring that new instructions can be satisfied without disturbing non-target content. Appendix~\ref{app:algorithm} provides the full executable loop and pseudocode.

\subsection{Architect: scene-grounded requirement compilation}
\label{sec:architect}

Forcing an LLM to satisfy a dense set of spatial constraints simultaneously often overloads its combinatorial reasoning, leading to dropped, hallucinated, or unverified constraints. The Architect addresses this by compiling $I_t$ into a structured \emph{requirement program} $\mathcal{B}_t$ whose units are typed, independently trackable, and schedulable. Rather than parsing language in isolation, it combines $I_t$ with the Scene Memory $\mathcal{G}_t$ and retrieved decomposition priors from $\mathcal{K}$, turning a holistic instruction into verifiable units and giving the downstream loop explicit targets for execution, retry, and preservation.

\paragraph{Requirement program structure.}
The requirement program $\mathcal{B}_t = (I_t,\; \textit{room\_type},\; \{r_1, \ldots, r_n\},\; \mathbf{g},\; \mathbf{s})$ consists of atomic requirements $r_i$, scene-aware constraints $\mathbf{g}$, and a style profile $\mathbf{s}$. Each requirement $r_i$ carries a type $\in \{\texttt{hard}, \texttt{edit}, \texttt{style}, \texttt{preserve}, \texttt{forbid}\}$, a priority, and an optional \emph{checkable predicate} $\phi_i$ for spatial and counting assertions; requirements with $\phi_i$ are verified deterministically, while those without receive model-based judgment (Appendix~\ref{app:brief_ir}). Here, \texttt{hard} denotes required construction constraints, \texttt{edit} denotes requested modifications, \texttt{style} denotes soft appearance preferences, \texttt{preserve} protects existing content, and \texttt{forbid} encodes negative constraints. By leveraging $\mathcal{G}_t$, the Architect resolves context-dependent references before handing requirements to the Sculptor; dependency completeness is further ensured by automatic entity injection (Appendix~\ref{app:enrichment}).

\paragraph{Requirement and feedback records.}
The structured requirement record used by the loop is
\[
    r_i = \bigl(\textit{type}_i,\; \textit{text}_i,\; \textit{priority}_i,\; \phi_i\bigr),
\]
where the optional predicate has the form
\[
    \phi_i = \bigl(\textit{predicate},\; \textit{subject},\; \textit{object},\; \textit{qualifiers}\bigr).
\]
The predicate field covers deterministically checkable spatial and counting assertions such as \texttt{count}, \texttt{exists}, \texttt{left\_of}, \texttt{right\_of}, \texttt{near}, \texttt{on\_top\_of}, and \texttt{against\_wall}. After each execution step, the Inspector returns feedback records
\[
    f_i = \bigl(\textit{status}_i,\; \textit{source}_i,\; \textit{matched\_ids}_i,\; \textit{failure\_mode}_i,\; \textit{repair\_hint}_i\bigr),
\]
where $\textit{status}_i \in \{\texttt{satisfied}, \texttt{unsatisfied}, \texttt{unknown}\}$. The $\textit{matched\_ids}_i$ field expands protection bindings when a requirement passes, while $\textit{failure\_mode}_i$ and $\textit{repair\_hint}_i$ provide targeted context for the next Sculptor step.

\paragraph{Skill memory $\mathcal{K}$: motif-guided decomposition.}
Compositional instructions such as ``bedside setup'' or ``dining area with chairs around the table'' often contain recurring structural motifs. Inspired by HSM~\cite{pun2026hsm}, which highlights reusable hierarchical motifs in indoor scenes, we treat these recurring decomposition patterns as reusable skills for controllable 3D scene authoring rather than as one-off parsing heuristics. Skill Memory stores such motif-level priors so that frequently recurring instruction patterns can be accumulated and reused, and later decomposed into separate, checkable requirements more reliably. During evaluation, $\mathcal{K}$ is frozen to prevent data leakage; outside evaluation, new motif cards can be added without any parameter updates (Appendix~\ref{app:enrichment}).

\subsection{Sculptor: requirement-grounded incremental execution}
\label{sec:sculptor}

Incremental execution creates a regression risk: later requirements may inadvertently undo content that has already been satisfied. The Sculptor addresses this by executing one requirement at a time on an explicitly instantiated scene state while maintaining protection over already-satisfied content, enabling preservation-aware local revision.

\paragraph{Scene-state grounding.}
Spatial tools require explicit geometric domains, such as support surfaces, wall domains, and free-floor regions that are not directly available from a raw object list. Before planning begins, the Sculptor constructs an initialized room state $\mathcal{R}_0$ from $\mathcal{S}_t$ and $\mathcal{G}_t$ to register these necessary structural anchors. This grounding step allows later declarative spatial intent to be mapped stably into concrete placements and transformations.

\paragraph{Requirement-grounded planning.}
To safely execute scheduled constraints, the system must track satisfaction status, pending targets, and objects demanding protection. Working Memory $\mathcal{W}_\tau$ records focus, satisfaction status, role bindings, and diagnostic history at each inner turn $\tau$ (full schema in Appendix~\ref{app:working_memory}). To prevent regressions, the Sculptor maintains two nested protection sets: the \textbf{protected set} $P_\tau$ forbids deletion or replacement of anchors, role-bound objects, and objects relied upon by satisfied requirements, while the \textbf{transform-protected set} $P^T_\tau \supseteq P_\tau$ additionally forbids moves, rotations, and resizes. After verification, these protection states are updated together with newly satisfied requirements, guaranteeing that later operations cannot compromise established progress.

\paragraph{Execution and local consistency pass.}
The Sculptor equips the planner with shared atomic authoring actions, encompassing both \emph{declarative} tools for expressing spatial relations and \emph{imperative} tools for direct object manipulation. The executor grounds declarative calls into geometric operations over the registered domains, allowing the planner to express spatial intent in a coordinate-free manner. Crucially, if the planner proposes an action that violates the protection sets $P_\tau$ or $P^T_\tau$, the executor preemptively blocks it before any scene modification occurs. 

After execution, an automated \emph{local consistency pass} resolves newly introduced collisions, support-placement offsets, and minor interpenetrations within the edited neighborhood. Because both preemptive blocking and local repair are constrained by $P_\tau$ and $P^T_\tau$, modifications remain highly localized, yielding a stable geometry for subsequent verification. More detailed tool definitions, fallback behaviors, and re-verification rules are deferred to Appendix~\ref{app:tools}.

\paragraph{Tool lowering and repair scope.}
Declarative calls such as \texttt{place\_on\_support}, \texttt{attach\_to\_wall}, and \texttt{add\_object\_with\_relation} are lowered into imperative placement calls only after the executor queries the relevant support surface, wall segment, or reference-object bounding box. The local consistency pass then inspects only the edited neighborhood: newly added or moved objects, their supports, and immediate collision partners. If a repair would require moving or deleting an object in $P_\tau$ or $P^T_\tau$, the repair is rejected and surfaced as an Inspector issue rather than silently damaging preserved content.

\subsection{Inspector: compositional verification and memory consolidation}
\label{sec:inspector}

Without explicit verification, the loop lacks reliable signals for deciding whether a specific constraint has been satisfied, which action should be retried, and when execution can stop. The Inspector addresses this by performing per-requirement verification and consolidating the result into persistent memory that provides the stopping, recovery, and continuation signals for subsequent execution.

\paragraph{Compositional verification.}
Per-requirement verification requires distinguishing between constraints that can be checked deterministically and those that require model-assisted judgment. We therefore combine two complementary mechanisms. If a requirement carries a checkable predicate $\phi_i$, a \emph{rule-based verifier} performs deterministic geometric checks such as counting matched objects, evaluating spatial predicates, and checking support or wall placement. For soft or perceptual aspects, a \emph{model-based feedback agent} judges the remaining soft or perceptual aspects from the repaired scene state and rendered views. Deterministic results take precedence whenever available, so model-based feedback complements rather than overrides verifiable constraints. The full feedback structure and reconciliation details are provided in Appendix~\ref{app:brief_ir}.

\paragraph{Memory consolidation and scheduling.}
Upon verification, the Inspector concurrently updates two memory layers. First, it updates Working Memory $\mathcal{W}_\tau$ with newly satisfied targets, expanding the nested protection sets ($P_\tau, P^T_\tau$) accordingly. Simultaneously, it consolidates the validated layout into Scene Memory $\mathcal{G}_\tau$, establishing a persistent structural graph (e.g., support hierarchies, zones) for future interactions. This dual-update ensures that later edits inherit a structured state rather than a raw object list. The scheduler then selects the next pending requirement, dynamically adjusting the execution order to prevent stalling on difficult constraints (Appendices~\ref{app:focus} and~\ref{app:scene_memory}).

\paragraph{Loop termination and global feasibility pass.}
The inner loop concludes once all scheduled requirements are verified or when further execution can no longer yield productive progress. Finally, the Inspector executes a \emph{preservation-aware global feasibility pass} that resolves any residual scene-level collisions and applies lightweight stability fixes. This yields a clean, structurally sound scene state ready for the next user instruction without compromising protected non-target content.

\section{AuthorBench: a compositional authoring benchmark}
\label{sec:benchmark}
\begin{figure}[t]
  \centering
  \includegraphics[width=\linewidth]{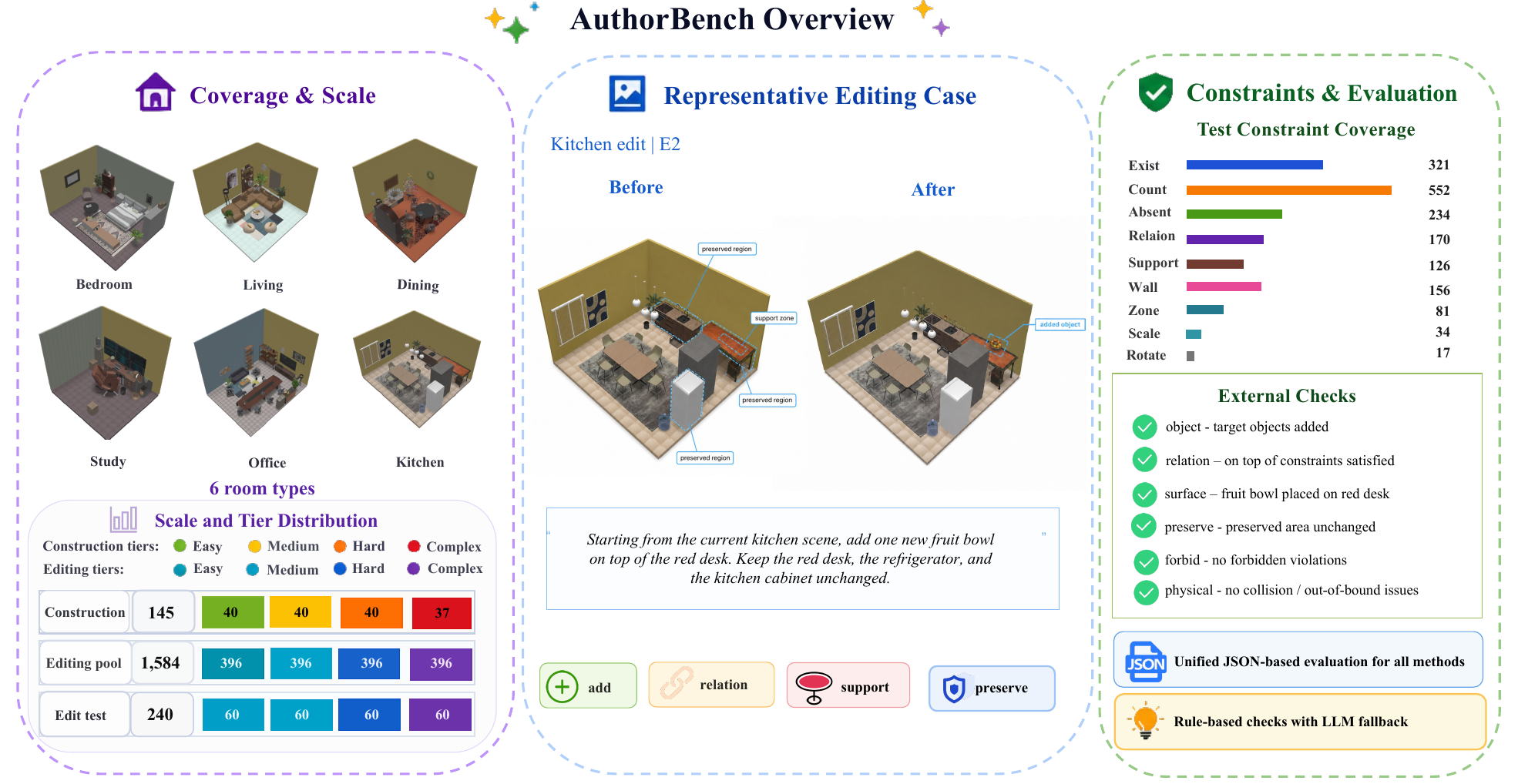}
  \vspace{0.3em}
  \caption{\textbf{AuthorBench overview.} AuthorBench unifies scene construction and preservation-aware editing under requirement-level evaluation. \textbf{(Left)} Dataset coverage across six room types and complexity tiers. \textbf{(Middle)} A representative editing case decomposing a user instruction into verifiable atomic constraints. \textbf{(Right)} The evaluation taxonomy, pairing each case with typed external checks.}
  \label{fig:benchmark_overview}
\end{figure}

Existing 3D scene-synthesis benchmarks mainly evaluate holistic scene quality, such as physical plausibility, visual realism, and semantic alignment~\cite{yang2025sceneweaver}, but they do not reveal whether each user constraint is satisfied and rarely test realistic local editing with preservation constraints. To fill this gap, we introduce \textbf{AuthorBench}. As illustrated in Figure~\ref{fig:benchmark_overview}, AuthorBench shifts the evaluation paradigm from holistic scene scoring to individual requirement verification. For fairness, each case pairs a natural-language input with external structured checks: programmable constraints are evaluated by deterministic predicates, while semantic constraints are evaluated by an independent visual-semantic judge, without relying on any method-internal verifier.

\paragraph{Subsets.}  AuthorBench contains 145 construction cases and 1584 single-instruction editing cases across six room types: bedroom, living room, dining room, study, office, and kitchen. Construction starts from an empty room; editing starts from source scenes drawn from Edit-As-Act~\cite{huang2026editasact} and Imaginarium~\cite{zhu2025imaginarium}, and specifically requires preservation constraints to test explicit local modifications rather than full-scene regeneration. All experiments use a fixed 240-case stratified editing split balanced by room type and Easy/Medium/Hard/Complex tiers; Appendix~\ref{app:authorbench} gives the sampling rules and tier definitions.

\paragraph{Annotation and metrics.} Annotation compiles natural-language requirements into typed requirement checks and then applies human review. Checks cover object, relation, surface, zone, preserve, and forbid types, enabling AuthorBench to evaluate goal completion, local preservation, and physical validity. Main experiments evaluate three dimensions: requirement satisfaction via \textit{Goal Fulfillment (GF)} and \textit{All-Goal} success; editing locality via \textit{Preservation Rate (PR)} and \textit{Unintended Change Rate (UCR)}; and physical validity via \textit{Out-Of-Bounds (OOB)} and \textit{Collisions (Col.)}.

\paragraph{Split construction and check coverage.}
The construction subset contains 145 cases across six room types: bedroom, living room, dining room, study, and office each contribute 25 cases, while kitchen contributes 20 cases. The editing benchmark contains a 1584-case instruction pool, balanced across E1 atomic, E2 grounded, E3 structured, and E4 compound edits with 396 cases per tier. The fixed 240-case editing test split is a deterministic stratified sample with 10 cases for every edit-complexity and room-type cell, yielding 40 editing cases per room type and 60 cases per edit tier. The released benchmark contains 1043 construction goal checks, 4140 editing-pool goal checks with 4752 preserve checks, and 648 goal checks with 720 preserve checks on the fixed editing test split. Detailed coverage is provided in Appendix~\ref{app:authorbench}.

\paragraph{External verifier inventory.}
AuthorBench scoring is intentionally separated from \textsc{MUSE}'s internal verifier. Deterministic checks are applied after scene export whenever the requirement can be expressed over object IDs, bounding boxes, support surfaces, wall domains, or zones. Residual semantic and perceptual constraints are judged by an independent visual-semantic judge that receives the same prompt, exported scene representation, and rendered views for all methods. The full predicate inventory is provided in Appendix~\ref{app:evaluation_protocol}.

\section{Experiments}
\label{sec:experiments}

Our evaluation assesses \textit{scene construction} (\S\ref{sec:exp_construction}), \textit{preservation-aware editing} (\S\ref{sec:exp_editing}), and \textit{additional analyses} covering human preference, component ablations, and downstream utility (\S\ref{sec:exp_analyses}). 

\textbf{Setup \& Baselines.} We evaluate released systems under the unified AuthorBench protocol (\S\ref{sec:benchmark}). Construction uses 145 cases starting from empty room shells; editing uses the fixed 240-case stratified split. We compare MUSE against LayoutGPT~\cite{feng2024layoutgpt}, Holodeck~\cite{yang2024holodeck}, LayoutVLM~\cite{sun2025layoutvlm}, ReSpace~\cite{bucher2025respace}, and SceneReVis~\cite{zhao2026scenerevis} for construction, and against LayoutGPT-E, Edit-As-Act~\cite{huang2026editasact}, SceneReVis, and ReSpace-E for editing. \textbf{Metrics} strictly follow \S\ref{sec:benchmark}; results are averaged across their respective evaluation subsets.

\textbf{Implementation details.}  
For comparable API-driven adapters, we use GPT-4o and convert outputs once into the common AuthorBench format for identical rendering and rule-based evaluation. We adapt LayoutGPT and ReSpace to editing as LayoutGPT-E and ReSpace-E: LayoutGPT-E conditions on the serialized source scene, while ReSpace-E uses its native structured-scene edit pipeline. Implementation, asset, and prompt details are in Appendices~\ref{app:baseline_impl}, \ref{app:asset_library}, and~\ref{app:prompt_templates}.

\textbf{Fairness protocol.}
For every baseline, we keep the released backbone, native prompting style, and stopping rule. The adapter standardizes only shared inputs and outputs: AuthorBench case definitions, room shells or source scenes, common assets when a conversion step is required, final rendering, and post-export evaluation. If a baseline emits a native layout or scene format, we convert it once into the unified AuthorBench scene JSON and score the resulting scene with the same AuthorBench external verifier used for all methods. Failed or partial runs are counted from the last emitted scene rather than manually repaired, so the reported comparison remains an end-to-end comparison under a shared benchmark protocol.

\subsection{Scene Construction}
\label{sec:exp_construction}

Table~\ref{tab:authorbench_construction} reports quantitative construction performance. MUSE is the only framework to achieve an \textit{All-Goal} success rate above 80\% (80.7), significantly outperforming the strongest one-shot baseline, LayoutGPT (37.9). When analyzing individual constraint families, the primary performance gap emerges in support-surface relations: MUSE fulfills 92.6\% of surface constraints, whereas baseline performance drops below 35.0\%. Furthermore, MUSE demonstrates consistent robustness across instruction difficulty. As evaluated by the complexity retention metric (Ret.), MUSE retains 96.6\% of its Easy-tier performance when evaluated on dense Complex-tier cases, while simultaneously avoiding physical errors (OOB 0.0, Col. 0.1).

\begin{table*}[t]
  \centering
  \caption{\textbf{Main construction results on AuthorBench.} We report requirement fulfillment, All-Goal success, complexity robustness (Ret.\ = Complex/Easy), and physical validity (OOB = out-of-boundary objects, Col.\ = colliding object pairs). Macro averages object, relation, surface, and zone fulfillment; Core averages object and relation fulfillment.}
  \label{tab:authorbench_construction}
  \scriptsize
  \setlength{\tabcolsep}{2.0pt}
  \renewcommand{\arraystretch}{1.04}
  \resizebox{\textwidth}{!}{%
  \begin{tabular}{lccccccccccccccc}
    \toprule
    \multirow{2}{*}{Method} & \multicolumn{7}{c}{Goal Fulfillment $\uparrow$} & \multirow{2}{*}{All-Goal$\uparrow$} & \multicolumn{5}{c}{Complexity Robustness $\uparrow$} & \multicolumn{2}{c}{Physical Validity $\downarrow$} \\
    \cmidrule(lr){2-8}\cmidrule(lr){10-14}\cmidrule(lr){15-16}
    & Obj. & Rel. & Surf. & Zone & Macro & Macro\textsuperscript{*} & Core & & Easy & Med. & Hard & Complex & Ret. & OOB & Col. \\
    \midrule
    LayoutGPT~\cite{feng2024layoutgpt} & 98.4 & 83.8 & 35.0 & 79.0 & 74.1 & 87.1 & 91.1 & 37.9 & 99.3 & 86.0 & 83.4 & 82.2 & 82.7 & 1.4 & 2.3 \\
    Holodeck~\cite{yang2024holodeck} & 83.3 & 57.6 & 21.5 & 30.9 & 48.3 & 57.3 & 70.5 & 14.5 & 84.7 & 73.7 & 67.9 & 54.1 & 63.9 & 0.0 & 0.0 \\
    LayoutVLM & 98.0 & 69.7 & 22.1 & 59.3 & 62.3 & 75.7 & 83.9 & 26.9 & 100.0 & 77.4 & 72.7 & 81.7 & 81.7 & 0.0 & 0.0 \\
    ReSpace~\cite{bucher2025respace} & 71.9 & 36.4 & 8.6 & 32.1 & 37.3 & 46.8 & 54.2 & 11.0 & 81.7 & 59.6 & 50.8 & 47.0 & 57.5 & 0.4 & 0.5 \\
    SceneReVis~\cite{zhao2026scenerevis} & 46.9 & 17.2 & 9.2 & 8.6 & 20.5 & 24.2 & 32.1 & 2.1 & 47.8 & 40.6 & 32.0 & 31.4 & 65.6 & 0.3 & 0.9 \\
    \rowcolor{black!6}\textsc{MUSE} & \textbf{99.3} & \textbf{93.9} & \textbf{92.6} & \textbf{82.7} & \textbf{92.1} & \textbf{92.0} & \textbf{96.6} & \textbf{80.7} & 99.3 & \textbf{97.9} & \textbf{94.9} & \textbf{95.9} & \textbf{96.6} & \textbf{0.0} & 0.1 \\
    \bottomrule
  \end{tabular}%
  }
\end{table*}

\begin{figure*}[t]
  \centering
  \includegraphics[width=0.96\textwidth]{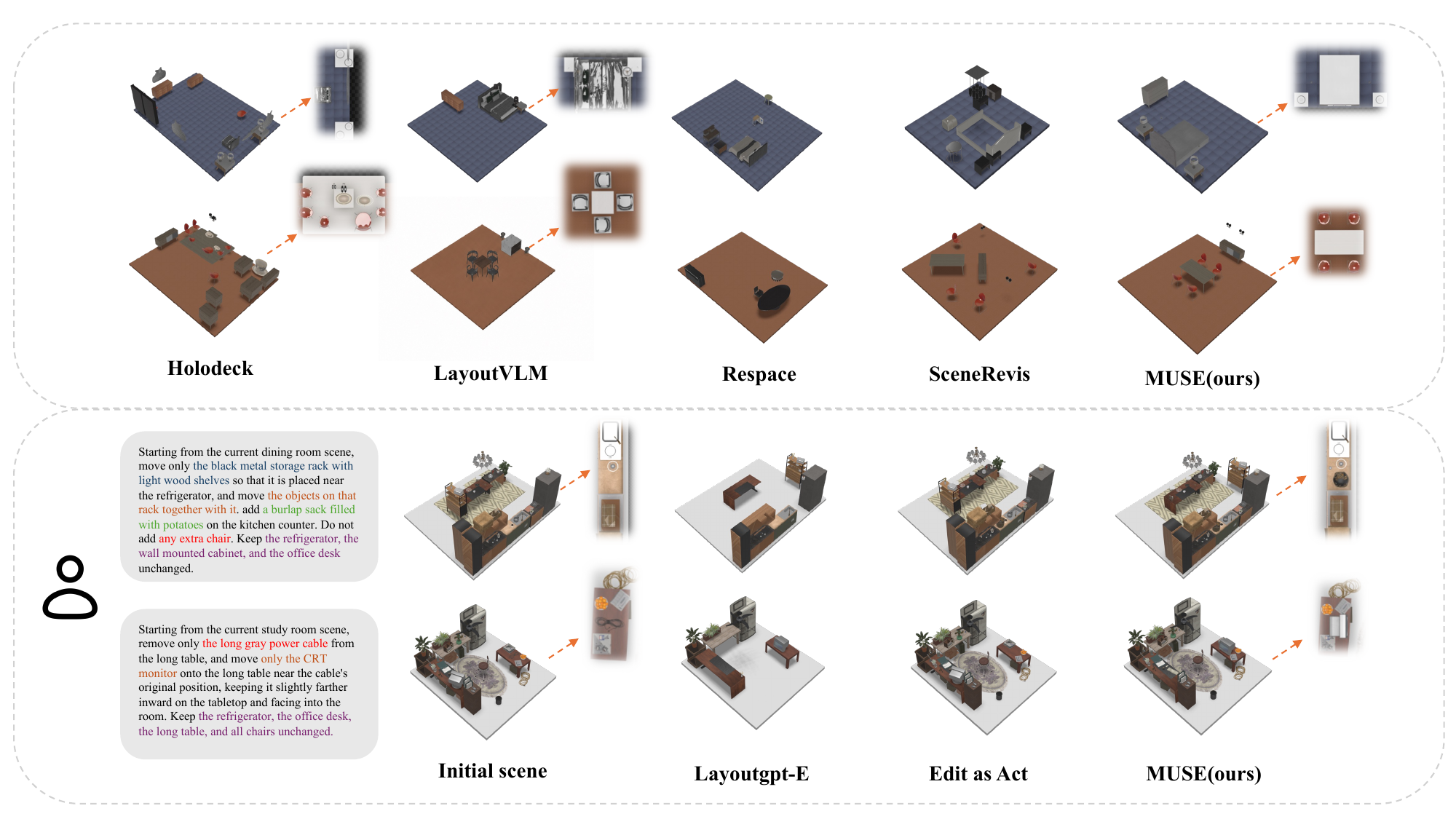}
  \vspace{0.3em}
  \caption{\textbf{Qualitative construction and editing comparisons.} Compared to baselines, MUSE generates more organized spatial layouts in scene construction, and performs precise local edits without breaking the original scene structure.}
  \label{fig:qualitative_main}
\end{figure*}

Qualitatively, Figure~\ref{fig:qualitative_main}(a) reflects these numerical differences. One-shot baselines often retrieve correct object categories but struggle with spatial dependencies: support objects are separated from their intended surfaces, and multi-object functional groups (e.g., dining sets) drift from the requested room-level arrangement. In contrast, MUSE preserves explicit global layout structures and grounds target objects precisely to their assigned surfaces.

\subsection{Preservation-Aware Scene Editing}
\label{sec:exp_editing}

A practical editor must satisfy new instructions without destroying unmentioned background. Table~\ref{tab:authorbench_editing} reports quantitative editing performance, revealing a clear completion-locality trade-off among baselines. Prompt-based regenerators (LayoutGPT-E) achieve high object fulfillment (86.0\%) but effectively overwrite the source scene (PR 25.1\%, UCR 95.7\%). Conversely, methods attempting local updates (Edit-As-Act, ReSpace-E, SceneReVis) execute only partial edits with redundant actions, yielding \textit{All-Goal} success below 12\%. MUSE mitigates this trade-off, achieving the highest task completion (Macro 69.0, All-Goal 49.6) while maintaining strict locality (PR 99.9\%, UCR 0.6\%) and near-zero edit-induced physical errors.

Appendix Figure~\ref{fig:editing_tradeoff} visualizes this trade-off. Baselines cluster either in high-damage regions (using rewriting as a shortcut) or low-completion regions (failing complex spatial logic). MUSE is positioned in the preferred top-left region, improving fulfillment without sacrificing stability. Figure~\ref{fig:qualitative_main}(b) qualitatively confirms this precise control. Given a compound instruction to move a rack, transfer its objects, and add a sack while protecting cabinets, LayoutGPT-E scrambles the layout, and Edit-As-Act fails the object transfer. MUSE confines modifications to targets, executing the coupled edit while largely preserving the protected context.

\par\begingroup\centering
  \captionof{table}{\textbf{Main editing results on AuthorBench.} We report fulfillment, All-Goal success, robustness, edit locality (PR/UCR), and edit-induced errors.}
  \label{tab:authorbench_editing}
  \scriptsize
  \setlength{\tabcolsep}{2.0pt}
  \renewcommand{\arraystretch}{1.12}
  \resizebox{\textwidth}{!}{%
  \begin{tabular}{lcccccccccccccc}
    \toprule
    \multirow{2}{*}{Method} & \multicolumn{4}{c}{Goal Fulfillment $\uparrow$} & \multirow{2}{*}{All-Goal$\uparrow$} & \multicolumn{5}{c}{Complexity Robustness $\uparrow$} & \multicolumn{2}{c}{Edit Locality} & \multicolumn{2}{c}{Physical Validity $\downarrow$} \\
    \cmidrule(lr){2-5}\cmidrule(lr){7-11}\cmidrule(lr){12-13}\cmidrule(lr){14-15}
    & Obj. & Rel. & Surf. & Macro & & Easy & Med. & Hard & Complex & Ret. & PR$\uparrow$ & UCR$\downarrow$ & OOB & Col. \\
    \midrule
    LayoutGPT-E~\cite{feng2024layoutgpt} & 86.0 & 21.1 & 19.3 & 42.2 & 36.2 & 86.7 & 59.2 & 61.0 & 67.2 & 77.5 & 25.1 & 95.7 & 1.8 & 5.2 \\
    Edit-As-Act~\cite{huang2026editasact} & 30.6 & 14.1 & 5.0 & 16.6 & 5.4 & 21.7 & 23.3 & 27.2 & 22.2 & 102.3 & 84.9 & 17.8 & 0.0 & 0.0 \\
    SceneReVis~\cite{zhao2026scenerevis} & 33.6 & 28.2 & 4.2 & 22.0 & 10.4 & 31.7 & 33.3 & 26.6 & 25.4 & 80.1 & 92.6 & 5.9 & 0.8 & 2.1 \\
    ReSpace-E~\cite{bucher2025respace} & 46.3 & 19.7 & 5.9 & 24.0 & 11.2 & 40.8 & 27.5 & 36.2 & 38.6 & 94.6 & 100.0 & 1.9 & 1.2 & 3.5 \\
    \rowcolor{black!6}\textsc{MUSE} & 76.0 & \textbf{47.9} & \textbf{83.2} & \textbf{69.0} & \textbf{49.6} & 73.3 & \textbf{71.7} & \textbf{68.6} & \textbf{80.2} & \textbf{109.4} & \textbf{99.9} & \textbf{0.6} & \textbf{0.1} & \textbf{0.3} \\
    \bottomrule
  \end{tabular}%
  }
\par\endgroup

\needspace{7\baselineskip}
\subsection{Additional Analyses}
\label{sec:exp_analyses}

\paragraph{Human evaluation.} To complement automated metrics, we invited 50 participants from diverse backgrounds to assess preservation-aware editing (details in Appendix~\ref{app:human_eval}). Table~\ref{tab:user_study} compares MUSE against Edit-As-Act, ReSpace-E, and SceneReVis. Raters evaluated instruction satisfaction, preservation, physical plausibility, and provided a 4-way overall preference. MUSE achieves the highest overall preference rate. Crucially, participants consistently penalized methods that introduced unnecessary background modifications. This alignment between human choices and our automated PR/UCR metrics indicates that MUSE's localized updates better match the evaluated user intent.

\noindent\textbf{\mbox{Component ablation.}}\enspace Removing per-step verification triggers a large drop in editing completion (Macro 69.0 $\rightarrow$ 18.5). Table~\ref{tab:ablation} isolates each framework component and shows that memory modules provide distinct causal benefits: protection sets govern locality (removal drops PR to 90.2\% and raises UCR to 9.7\%), while Scene Memory maintains structural grounding. Notably, Skill Memory is critical for complex compositional reasoning; removing it disproportionately degrades Complex-tier performance, lowering the overall editing Macro to 61.4.

\par\smallskip\begingroup\centering
  \centering
  \captionof{table}{\textbf{Component ablation.} Rows remove one component from \textsc{MUSE}; metrics use the fixed 240-case split.}
  \label{tab:ablation}
  \scriptsize
  \setlength{\tabcolsep}{2.4pt}
  \renewcommand{\arraystretch}{0.98}
  \resizebox{\textwidth}{!}{%
  \begin{tabular}{lcccccccc}
    \toprule
    & \multicolumn{4}{c}{Construction} & \multicolumn{3}{c}{Editing} & \multicolumn{1}{c}{Phys.} \\
    \cmidrule(lr){2-5}\cmidrule(lr){6-8}\cmidrule(lr){9-9}
    Variant & Macro-GF$\uparrow$ & Hard$\uparrow$ & Complex$\uparrow$ & HCS$\uparrow$ & Macro$\uparrow$ & PR$\uparrow$ & UCR$\downarrow$ & Col.$\downarrow$ \\
    \midrule
    \textsc{MUSE} (full)                      & \textbf{92.1} & \textbf{95.0} & \textbf{95.9} & \textbf{95.3} & \textbf{69.0} & \textbf{99.9} & \textbf{0.6} & \textbf{0.3} \\
    \quad w/o focus scheduler                 & 90.9 & 92.6 & 93.7 & 93.8 & 66.4 & 97.5 & 2.5 & 0.4 \\
    \quad w/o protection sets                 & 91.0 & 93.2 & 94.4 & 94.1 & 65.3 & 90.2 & 9.7 & 0.5 \\
    \quad w/o Scene Memory                    & 89.4 & 91.2 & 92.1 & 92.4 & 63.2 & 95.3 & 4.2 & 0.5 \\
    \quad w/o Skill Memory                    & 88.7 & 92.4 & 89.6 & 91.8 & 61.4 & 96.5 & 3.4 & 0.4 \\
    \quad single-pass                         & 14.5 & 26.8 & 25.5 & 34.4 & 18.5 & 76.2 & 19.8 & 0.6 \\
    \bottomrule
  \end{tabular}%
  }
\par\endgroup

\paragraph{Downstream utility.}
Table~\ref{tab:navigation_curriculum} evaluates 100 three-stage navigation curricula, producing 300 scenes per method. While both incremental MUSE and stage-wise regeneration produce navigable final scenes (high reachable ratios, zero collisions), MUSE yields substantially stronger spatial continuity. It achieves higher difficulty monotonicity (0.850 vs.\ 0.394) and free-space preservation (0.978 vs.\ 0.649). This confirms that incremental authoring provides not merely static navigability, but a stable, predictable spatial substrate useful for navigation-oriented embodied-AI settings.

\par\smallskip\begingroup\centering
  \centering
  \newlength{\mainpairedtableheight}
  \setlength{\mainpairedtableheight}{0.95in}
  \begin{minipage}[t][\mainpairedtableheight][t]{0.45\textwidth}
    \centering
    \captionof{table}{\textbf{Human evaluation.} Mean ratings (1--5) and 4-way overall preference (\%).}
    \label{tab:user_study}
    \scriptsize
    \setlength{\tabcolsep}{3pt}
    \renewcommand{\arraystretch}{0.96}
    \vspace{0.3em}
    \vfill
    \begin{tabular}{lcccc}
      \toprule
      Method & Satis.$\uparrow$ & Preserv.$\uparrow$ & Plaus.$\uparrow$ & Pref.$\uparrow$ \\
      \midrule
      Edit-As-Act & 2.5 & 4.1 & 4.3 & 16.2\% \\
      SceneReVis  & 3.2 & 3.5 & 3.8 & 11.0\% \\
      ReSpace-E   & 3.6 & 3.8 & 3.5 & 9.4\% \\
      \textsc{MUSE} (Ours) & \textbf{4.5} & \textbf{4.8} & \textbf{4.6} & \textbf{63.4\%} \\
      \bottomrule
    \end{tabular}
  \end{minipage}\hfill
  \begin{minipage}[t][\mainpairedtableheight][t]{0.53\textwidth}
    \centering
    \captionof{table}{\textbf{Downstream curricula.} 100-case / 300-scene navigation with Mono./FS-IoU consistency metrics.}
    \label{tab:navigation_curriculum}
    \scriptsize
    \setlength{\tabcolsep}{2.0pt}
    \renewcommand{\arraystretch}{0.96}
    \vspace{0.3em}
    \vfill
    \begin{tabular}{@{}lcccccc@{}}
      \toprule
      Method & Mono.$\uparrow$ & FS-IoU$\uparrow$ & Walk@3 & Reach@3 & Path@3 & Col./OOB \\
      \midrule
      Stage-wise  & 0.394 & 0.649 & \textbf{0.878} & \textbf{0.999} & 2.183 & 0.0/0.0 \\
      \textsc{MUSE} & \textbf{0.850} & \textbf{0.978} & 0.868 & 0.992 & \textbf{1.736} & 0.0/0.0 \\
      \bottomrule
    \end{tabular}
  \end{minipage}
\par\endgroup

\vspace{1.0em}

\section{Conclusion}
\label{sec:conclusion}

In this paper, we formulated controllable 3D scene authoring as incremental requirement satisfaction and introduced \textsc{MUSE}, a memory-grounded multi-agent framework for requirement compilation, iterative execution, and step-wise verification. Together with AuthorBench, a requirement-level benchmark for construction and preservation-aware editing, our experiments show that \textsc{MUSE} outperforms one-shot baselines in dense construction and mitigates the editing completion-locality trade-off, achieving strong task completion with near-perfect structural stability (PR 99.9\%, UCR 0.6\%). Human evaluations and downstream navigation-proxy tests further support its alignment with user intent and spatial stability.

\textbf{\small Limitation.} \textsc{MUSE} and AuthorBench are currently limited to indoor scenes and single-instruction edits, and \textsc{MUSE} still relies on repeated LLM-guided verification. Future work will extend the domain scope, distill verification into efficient models, and support long-horizon multi-turn authoring. These results point toward a shift from single-shot holistic generation to verifiable, requirement-level 3D scene progression.

\clearpage
\appendix

\section*{Appendix}

The appendix is organized into five parts. Appendix~A gives method and algorithm details. Appendix~B documents AuthorBench and the evaluation protocol. Appendix~C provides additional experimental results. Appendix~D covers implementation, assets, prompts, and the frontend. Appendix~E discusses limitations and broader impacts.

\section{Method and Algorithm Details}
\label{app:method_details}

This part expands the closed-loop authoring process, structured state, scheduling logic, and execution machinery that are summarized in the main paper.

\subsection{Closed-loop execution algorithm}
\label{app:algorithm}

This section expands the method with the supporting implementation details that are abbreviated in the main text. Table~\ref{tab:arxiv_memory_state} summarizes the structured state components referenced throughout the algorithm.

\begin{table}[t]
\centering
\caption{\textbf{Structured state used by MUSE.} The main fields support progress tracking, preservation, and later-turn grounding. Working Memory is per-input; Scene Memory persists across interactions.}
\label{tab:arxiv_memory_state}
\footnotesize
\setlength{\tabcolsep}{3pt}
\renewcommand{\arraystretch}{1.08}
\begin{tabularx}{\linewidth}{@{}p{0.22\linewidth}p{0.25\linewidth}Y@{}}
\toprule
\textbf{State} & \textbf{Stored fields} & \textbf{Role in the authoring loop} \\
\midrule
Working focus & Current requirement, pending/done partition, diagnostic issues & Selects the next local action and detects stalls \\
Protection state & Target set $T$, protected set $P$, transform-protected set $P^T$ & Blocks deletion, replacement, movement, rotation, and resizing of protected content \\
Role bindings & Requirement-to-object matches and semantic role links & Grounds later references and expands protection after verification \\
Action history & Tool calls, blocked actions, degradations, repair attempts & Explains failures and conditions the next planning step \\
Scene graph & Stable object IDs, support edges, relation edges, wall anchors & Maintains editable structure across turns \\
Spatial domains & Room shell, wall domains, support surfaces, zones, free-floor regions & Grounds declarative placement and constrains local repair \\
\bottomrule
\end{tabularx}
\end{table}

\subsection{Requirement program and feedback records}
\label{app:brief_ir}

The requirement program is defined as:
\[
    \mathcal{B}_t = \bigl(I_t,\; \textit{room\_type},\; \{r_1, \ldots, r_n\},\; \mathbf{g},\; \mathbf{s}\bigr),
\]
where $\mathbf{g}$ denotes scene-aware constraints grounded in the current room context, and $\mathbf{s}$ a style profile. Each requirement $r_i$ is a structured record:
\[
    r_i = \bigl(\textit{type}_i,\; \textit{text}_i,\; \textit{priority}_i,\; \phi_i\bigr),
\]
with $\textit{type}_i \in \{\texttt{hard}, \texttt{edit}, \texttt{style}, \texttt{preserve}, \texttt{forbid}\}$. The optional checkable predicate $\phi_i$ has the form:
\[
    \phi_i = \bigl(\textit{predicate},\; \textit{subject},\; \textit{object},\; \textit{qualifiers}\bigr),
\]
where \textit{predicate} $\in$ \{\texttt{count}, \texttt{exists}, \texttt{left\_of}, \texttt{right\_of}, \texttt{near}, \texttt{on\_top\_of}, \texttt{against\_wall}, \ldots\} defines a deterministically checkable spatial or counting assertion, and \textit{subject}/\textit{object} are typed references carrying \texttt{semantic\_type}, \texttt{canonical\_type}, and \texttt{role} fields for robust matching.

The Inspector returns a structured feedback record
\[
    \mathcal{F}_\tau = \{f_i\}_{i=1}^{n}, \qquad
    f_i = \bigl(\textit{status}_i,\; \textit{source}_i,\; \textit{matched\_ids}_i,\; \textit{failure\_mode}_i,\; \textit{repair\_hint}_i\bigr),
\]
where $\textit{status}_i \in \{\texttt{satisfied}, \texttt{unsatisfied}, \texttt{unknown}\}$ and $\textit{source}_i$ records whether the decision came from a rule-based predicate, model-assisted judgment, or both. When a rule-based predicate is available, its result takes precedence for the corresponding hard spatial or counting assertion; model-assisted feedback is used for residual style, perceptual, or open-ended semantic aspects. The reconciled record updates Working Memory by moving satisfied requirements to $\textit{done}_\tau$, attaching matched object IDs to protection bindings, and recording failure modes or repair hints for the next planning turn.

\subsection{Working Memory and protection state}
\label{app:working_memory}

Working Memory $\mathcal{W}_\tau$ at inner turn $\tau$ maintains the full execution state:
\[
    \mathcal{W}_\tau = \bigl(\textit{focus}_\tau,\; \textit{done}_\tau,\; \textit{pending}_\tau,\; \Pi_\tau,\; \textit{bindings}_\tau,\; \textit{actions}_\tau,\; \textit{issues}_\tau\bigr),
\]
where $\textit{focus}_\tau$ is the current requirement, $\textit{done}_\tau / \textit{pending}_\tau$ partition requirements by satisfaction status, $\Pi_\tau = (T_\tau, P_\tau, P^T_\tau)$ is the protection state (target, protected, transform-protected sets), $\textit{bindings}_\tau$ maps semantic roles to object IDs, $\textit{actions}_\tau$ logs tool calls and degradations, and $\textit{issues}_\tau$ stores diagnostic messages from verification.

\subsection{Scene Memory organization}
\label{app:scene_memory}

Scene Memory $\mathcal{G}_t$ stores the persistent spatial state: room shell, wall domains, support surfaces, free-floor regions, functional zones, stable object IDs, and relation edges (support, adjacency, wall anchoring, zone membership). After each Inspector update, satisfied requirement bindings are reflected in this graph so that later turns can resolve references against the current scene. Zone-level bookkeeping groups objects for locality-aware scheduling and preservation. Scene Memory persists across turns, while Working Memory is reinitialized per input.

\subsection{Cross-turn memory inheritance and conflict handling}
\label{app:cross_turn}

Working Memory $\mathcal{W}_\tau$ is an intra-turn structure and does not persist across user turns. At the beginning of a new instruction, MUSE reinitializes working memory from the current scene state and recompiles the new requirement program. What persists is scene memory $\mathcal{G}_{t+1}$: stable object IDs, the updated room graph, support and wall domains, zone annotations, and any grounded state needed to interpret the next instruction.

This design means that contradictory later edits are handled by recompiling against the current scene rather than enforcing a global monotonicity rule. For example, if one turn adds a sofa near the window and the next turn requests removing that sofa, the second turn simply targets the grounded sofa instance in the updated graph and removes it. The preservation guarantee therefore applies within a single authoring turn through the protection sets and verifier loop; across turns, MUSE follows the latest user instruction while retaining only the scene state needed for grounded continuation.

\subsection{Focus scheduling and stall recovery}
\label{app:focus}

The focus scheduler selects the next requirement using a multi-signal scoring function. Key signals include: verifiability bonus (+100 for rule-based checks), constraint kind priority (entity +35, edit +45, placement +10, layout $-$15), scope locality (+30--45 for same-zone continuation), priority weight (up to +20), and self-repetition penalty ($-$50 to avoid stalls). Candidates are ranked by score; the top candidate is selected if its lead exceeds 25 points, otherwise top-3 are sampled proportionally. Requirements with unmet dependencies are ineligible. The scheduler tracks ineffective focus counts; after $k{=}3$ consecutive stalled turns, a requirement is temporarily deprioritized.

\subsection{Tool hierarchy and consistency passes}
\label{app:tools}

The Sculptor's action vocabulary has two layers. \textbf{Declarative tools} express spatial intent without coordinates: \texttt{add\_object\_with\_relation}, \texttt{place\_on\_support}, \texttt{place\_under\_support}, \texttt{attach\_to\_wall}, \texttt{reposition\_object\_by\_relation}. \textbf{Imperative tools} specify exact parameters: \texttt{add\_object}, \texttt{remove\_object}, \texttt{move\_object}, \texttt{rotate\_object}, \texttt{scale\_object}, \texttt{replace\_object}, \texttt{terminate}.

The executor compiles declarative calls into imperative operations through domain-specific geometric reasoning over support surfaces, wall domains, and relational offsets. When compilation fails, graceful degradation applies (e.g., support fallback to nearby floor placement), logged as \texttt{ToolDegradation} events.

The \textbf{local consistency pass} inspects only the edited neighborhood (newly added/moved objects, supports, collision partners) and resolves support-height offsets, wall flushness errors, and small collisions with minimal adjustment. Objects in $P_\tau$ and $P^T_\tau$ are never moved or deleted; if repair would violate protection, the issue is surfaced to the Inspector. The \textbf{global feasibility pass} (after loop termination) checks the full scene for residual collisions and out-of-boundary placements, prioritizing protected content preservation.

\section{AuthorBench and Evaluation Protocol}
\label{app:evaluation_protocol}

This part specifies the benchmark composition, metrics, external verifier, baseline adapters, and human evaluation protocol used for all reported results.

\begin{table}[t]
\centering
\caption{\textbf{AuthorBench coverage.} The fixed editing test split preserves the main operation families from the 1584-case pool while enforcing room-level and edit-tier balance.}
\label{tab:arxiv_authorbench_coverage}
\footnotesize
\setlength{\tabcolsep}{4pt}
\renewcommand{\arraystretch}{1.08}
\begin{minipage}[t]{0.48\linewidth}
\centering
\begin{tabular}{@{}lrr@{}}
\toprule
\multicolumn{3}{@{}l@{}}{\textbf{(a) Edit design}} \\
\midrule
\textbf{Statistic} & \textbf{Pool} & \textbf{Test} \\
\midrule
E1 atomic & 396 & 60 \\
E2 grounded & 396 & 60 \\
E3 structured & 396 & 60 \\
E4 compound & 396 & 60 \\
Imaginarium & 1488 & 193 \\
Edit-As-Act & 96 & 47 \\
\bottomrule
\end{tabular}
\end{minipage}\hfill
\begin{minipage}[t]{0.48\linewidth}
\centering
\begin{tabular}{@{}lrr@{}}
\toprule
\multicolumn{3}{@{}l@{}}{\textbf{(b) Scene and target burden}} \\
\midrule
\textbf{Statistic} & \textbf{Pool} & \textbf{Test} \\
\midrule
S1 simple & 300 & 35 \\
S2 medium & 1008 & 135 \\
S3 dense & 192 & 41 \\
S4 extreme & 84 & 29 \\
Q0 single target & 1068 & 161 \\
Q1 two--three targets & 374 & 57 \\
Q2 four-plus targets & 142 & 22 \\
\bottomrule
\end{tabular}
\end{minipage}
\end{table}

\subsection{Metric definitions}
\label{app:metrics}

\paragraph{Goal fulfillment.} For a case with annotated goal checks $\mathcal{G}$, GF is the fraction of satisfied checks:
\[
    \mathrm{GF} = \frac{1}{|\mathcal{G}|}\sum_{g \in \mathcal{G}}\mathbf{1}[g\ \mathrm{satisfied}].
\]
We group checks into constraint families: object constraints (existence, absence, count, scale, and rotation), relation constraints, surface constraints (support and wall placement), and zone constraints (functional zones and clear-area requirements). The benchmark annotation covers the same four families for both construction and editing. Macro-GF is the unweighted mean of the reported family GF values. In the main construction table, we additionally report Macro\textsuperscript{*}, the mean of object/relation/zone GF, and Core, the mean of object/relation GF, to separate support-surface difficulty from the rest of compositional satisfaction. In the main editing table, we report Obj./Rel./Surf. and compute Macro over these reported families only, omitting Zone for readability in the current comparison. All-Goal success is the fraction of cases where every annotated goal check is satisfied.

\paragraph{Editing locality.} For editing cases, preservation rate (PR) measures the fraction of protected checks that remain valid:
\[
    \mathrm{PR} = \frac{1}{|\mathcal{P}|}\sum_{p \in \mathcal{P}}\mathbf{1}[p\ \mathrm{preserved}].
\]
Unintended change rate (UCR) measures accidental modification of non-target initial objects:
\[
    \mathrm{UCR} =
    \frac{\#\{\mathrm{changed\ non\mbox{-}target\ objects}\}}
    {\#\{\mathrm{non\mbox{-}target\ objects}\}}.
\]

\paragraph{Physical validity.} OOB (out-of-boundary) and Col.\ (collisions) are the average number of out-of-boundary objects and colliding object pairs, computed from the final scene geometry. For construction, these measure the physical validity of the generated scene. For editing, these measure new physical issues introduced by the edit operation, assuming the source scene is physically valid. HCS is a validity-weighted hard-constraint score: the pass rate over existence, absence, count, relation, support, and wall checks, multiplied by $1/(1+\mathrm{OOB}+\mathrm{Col.})$. It is used only for ablations and diagnostics.

\subsection{AuthorBench construction and editing details}
\label{app:authorbench}

This section expands the dataset description in \S\ref{sec:benchmark}.

\paragraph{Subset sizes and tier balance.} The construction subset contains 145 cases across six room types: bedroom, living room, dining room, study, and office (25 cases each), and kitchen (20 cases). The editing benchmark consists of a 1584-case instruction pool plus the fixed 240-case test split used throughout this paper. The full editing pool is balanced by edit complexity, with 396 cases in each of E1 atomic, E2 grounded, E3 structured, and E4 compound. The test split is a deterministic stratified sample with 10 cases for each (edit-complexity, room-type) cell, giving 40 editing cases per room type and 60 cases per complexity tier. The released benchmark contains 1043 construction goal checks, 4140 editing-pool goal checks with 4752 preserve checks, and 648 goal checks with 720 preserve checks on the fixed editing test split.

\begin{figure}[t]
    \centering
    \includegraphics[width=\linewidth]{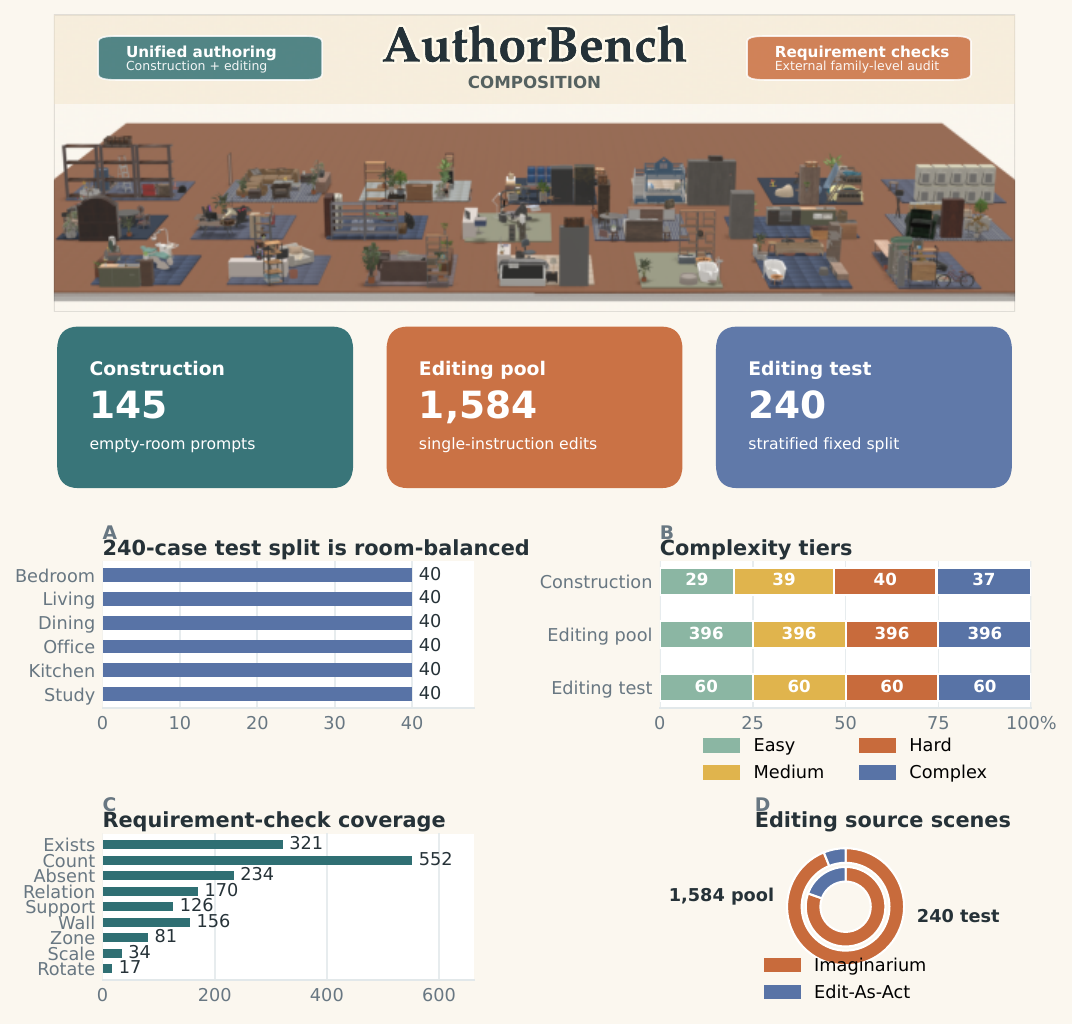}
    \caption{\textbf{AuthorBench composition.} The benchmark combines 145 construction cases, a 1584-case editing pool, and a fixed 240-case editing test split. The split is balanced by room type and edit complexity while retaining coverage across source-scene families and check types.}
    \label{fig:authorbench_composition}
\end{figure}

\begin{table}[t]
\centering
\caption{\textbf{Editing pool and test-split coverage.} (a) summarizes edit tiers and source families; (b) summarizes source-scene complexity and target burden. The fixed test split preserves the main operation families from the 1584-case pool while enforcing room-level balance.}
\label{tab:authorbench_editing_coverage}
\footnotesize
\setlength{\tabcolsep}{4pt}
\renewcommand{\arraystretch}{1.08}
\begin{minipage}[t]{0.48\linewidth}
\centering
\begin{tabular}{@{}lrr@{}}
\toprule
\multicolumn{3}{@{}l@{}}{\textbf{(a) Edit design}} \\
\midrule
\textbf{Statistic} & \textbf{Pool} & \textbf{Test} \\
\midrule
E1 atomic & 396 & 60 \\
E2 grounded & 396 & 60 \\
E3 structured & 396 & 60 \\
E4 compound & 396 & 60 \\
Imaginarium & 1488 & 193 \\
Edit-As-Act & 96 & 47 \\
\bottomrule
\end{tabular}
\end{minipage}\hfill
\begin{minipage}[t]{0.48\linewidth}
\centering
\begin{tabular}{@{}lrr@{}}
\toprule
\multicolumn{3}{@{}l@{}}{\textbf{(b) Scene and target burden}} \\
\midrule
\textbf{Statistic} & \textbf{Pool} & \textbf{Test} \\
\midrule
S1 simple & 300 & 35 \\
S2 medium & 1008 & 135 \\
S3 dense & 192 & 41 \\
S4 extreme & 84 & 29 \\
Q0 single target & 1068 & 161 \\
Q1 two--three targets & 374 & 57 \\
Q2 four-plus targets & 142 & 22 \\
\bottomrule
\end{tabular}
\end{minipage}
\end{table}

\paragraph{Three-stage construction pipeline.} Each case is produced through:
\begin{itemize}[nosep,leftmargin=*]
    \item \emph{(i) Skeleton drafting.} Per room type, we instantiate skeleton templates parameterized by required object inventory, allowed/forbidden categories, anchor surfaces, wall references, and functional-zone composition. Templates are sampled across the four tiers so that tier statistics are balanced within each room type.
    \item \emph{(ii) Constraint compilation.} Each skeleton is expanded into a natural-language prompt and an external constraint set. Every constraint carries one of six types (object, relation, surface, zone, preserve, forbid) and, where feasible, a machine-checkable predicate (support-on, wall-anchored-to, count, negative-existence, free-floor area; see Appendix~\ref{app:verifier}).
    \item \emph{(iii) Human review.} Two annotators independently verify that the prompt, source scene (for editing), and predicate set are mutually consistent, and that the case is realizable. Disagreements are revised or discarded.
\end{itemize}

\paragraph{Tier definitions.} Easy cases mainly test object inventory, count, and negative constraints. Medium cases add spatial relations and wall placement. Hard cases introduce support-surface reasoning and functional-zone composition. Complex cases mix multiple constraint families in dense compound prompts or edits, increasing the number of interacting goals and protected objects.

\paragraph{Editing source scenes.} Editing source scenes are drawn from Edit-As-Act~\cite{huang2026editasact} and Imaginarium~\cite{zhu2025imaginarium}, then normalized into the AuthorBench scene schema. Normalization assigns stable object identifiers, canonical category names, simple geometry, support surfaces, wall anchors, and functional zones, so target grounding (``the desk on the east wall'') is unambiguous and protection checks can be verified against original object IDs. The same normalized source scene is provided to every method for a given editing case, so editing scores reflect local modification and preservation rather than differences in upstream scene generation. The benchmark-side external checks are authored from the case skeletons plus human review, rather than copied from \textsc{MUSE}'s compiled requirement program, and are not exposed to any method as gold test-time supervision.

\paragraph{Case coverage.} Construction cases span object inventory, count, and negative constraints (Easy), spatial and wall-placement (Medium), support-surface and zone composition (Hard), and dense compound prompts mixing all families (Complex). Editing cases include atomic edits (add/move/remove/replace) and compound instructions, with boundary cases that target support-surface management, wall-mounted objects, target grounding, and controlled local modification.

\paragraph{Split balancing and reporting.} The fixed construction and editing splits are balanced jointly by room type and difficulty tier, and the editing split is additionally balanced by edit family. These distributions are used only to construct a stable and reviewer-auditable benchmark protocol; they are not exposed to the method at test time. The predicate inventory used by the external checks is listed separately in Appendix~\ref{app:verifier}.

\paragraph{Independent semantic judging.} Deterministic predicates are used whenever a requirement can be expressed over the exported scene graph or geometry. Residual semantic or perceptual constraints are evaluated by an independent judge that receives the original requirement, the exported scene representation, and rendered views under the same prompt and protocol for all methods. The judge is reserved for non-programmable constraints and never overrides deterministic predicate outcomes, preserving the separation between AuthorBench evaluation and any method-internal verifier. This means AuthorBench is benchmark-side and method-agnostic: the same checking protocol is applied to all exported scenes regardless of whether a method internally uses requirement programs, action plans, or holistic generation.

\subsection{External verifier predicate catalog}
\label{app:verifier}

The rule-based verifier implements deterministic geometric checks over typed \texttt{VerifierConstraint} records. It covers: \textbf{entity predicates} (\texttt{count}, \texttt{exists}); \textbf{relative placement} (\texttt{left\_of}, \texttt{right\_of}, \texttt{front\_of}, \texttt{behind}, \texttt{near}, \texttt{far}, \texttt{parallel\_to}); \textbf{support and wall checks} (\texttt{on\_top\_of}, \texttt{under\_support}, \texttt{against\_wall}); \textbf{zone, preserve, forbid}, and \textbf{physical validity} checks. Table~\ref{tab:verifier_family_inventory} summarizes these families. For each predicate, the verifier produces structured diagnostics: \texttt{failure\_mode}, \texttt{repair\_hint}, \texttt{matched\_ids}, and quantitative \texttt{metrics}.

\begin{table}[t]
\centering
\caption{\textbf{AuthorBench predicate inventory by requirement family.} External checks are evaluated after scene export and are independent of \textsc{MUSE}'s internal verifier.}
\label{tab:verifier_family_inventory}
\footnotesize
\setlength{\tabcolsep}{3pt}
\renewcommand{\arraystretch}{1.05}
\begin{tabularx}{\linewidth}{@{}p{0.14\linewidth}p{0.42\linewidth}Y@{}}
\toprule
\textbf{Family} & \textbf{Predicates / checks} & \textbf{Typical failure modes} \\
\midrule
Object & \texttt{exists}, \texttt{count}, \texttt{negative-existence}, category match & Missing object, extra forbidden object, wrong category \\
Relation & \texttt{left\_of}, \texttt{right\_of}, \texttt{front\_of}, \texttt{behind}, \texttt{near}, \texttt{far}, \texttt{parallel\_to} & Wrong side, too far/near, orientation mismatch \\
Surface & \texttt{on\_top\_of}, \texttt{under\_support}, \texttt{wall-anchored-to}, \texttt{against\_wall} & Unsupported object, wall offset, wrong support \\
Zone & \texttt{inside\_zone}, \texttt{zone\_contains}, \texttt{free-floor-area} & Object outside zone, cluttered zone, insufficient clear area \\
Preserve & \texttt{preserve\_existence}, \texttt{preserve\_pose}, \texttt{preserve\_relation}, \texttt{preserve\_support} & Deleted non-target, moved anchor, broken relation \\
Forbid & \texttt{forbid\_category}, \texttt{forbid\_region}, \texttt{forbid\_relation} & Forbidden object, forbidden placement \\
Physical & OOB, collision pairs, support-height consistency & Out-of-room, interpenetration, floating/sunken \\
\bottomrule
\end{tabularx}
\end{table}

\subsection{Verifier coverage and fallback judgment}
\label{app:verifier_coverage}

The external AuthorBench checks are independent from MUSE's internal verifier. MUSE uses requirement-level verification inside the loop to decide what to do next, but all reported paper metrics are computed from the external benchmark checks after scene export. This separation prevents the system from being rewarded for satisfying only its own internal judgment interface.

In practice, the rule-based verifier covers grounded count, existence, relation, support, wall, and zone-like checks whenever the requirement can be expressed as a typed predicate over the scene graph. Model-assisted judgment is reserved for residual style or open-ended semantic feedback that is difficult to encode as a deterministic geometric rule. When the internal verifier cannot certify a requirement, it may still provide guidance to the planner, but the benchmark score always comes solely from the external checks defined by AuthorBench. The external predicate catalog is method-agnostic: any baseline or future method that exports an evaluable scene in the unified AuthorBench schema can be scored by the same post-export checker, without requiring it to share \textsc{MUSE}'s internal decomposition or verifier interface.

\subsection{Baseline adapters and fairness protocol}
\label{app:baseline_impl}

We keep each baseline's released backbone, native prompting style, and stopping rule, and standardize what is shared: the AuthorBench case definitions, room shells or source scenes, the common asset pool when conversion is required, and the final rendering and evaluation stack. When a baseline emits a native layout or scene format, we convert it once into the unified AuthorBench scene JSON and score it with the same external verifier used for all methods. Failed or partial runs are counted from the method's last emitted scene rather than manually repaired. Table~\ref{tab:baseline_interfaces} summarizes the adapter interfaces.

\begin{table}[t]
\centering
\caption{\textbf{Baseline interfaces and AuthorBench adapters.} Each baseline keeps its released backbone and native control interface; the adapter standardizes the input scene format and final exported scene for external AuthorBench evaluation.}
\label{tab:baseline_interfaces}
\footnotesize
\setlength{\tabcolsep}{3pt}
\renewcommand{\arraystretch}{1.05}
\begin{tabularx}{\linewidth}{@{}p{0.20\linewidth}p{0.28\linewidth}Y@{}}
\toprule
\textbf{Baseline} & \textbf{Native interface} & \textbf{AuthorBench adapter} \\
\midrule
LayoutGPT / LayoutGPT-E & One-shot text-to-layout prediction & Prompt + room/source serialization to scene JSON \\
LayoutVLM & Construction-only layout optimization & Empty shell + asset metadata to scene JSON \\
Holodeck & Construction-only generation & Empty shell + prompt to scene JSON \\
ReSpace / ReSpace-E & Structured scene representation & Convert SSR scene to AuthorBench JSON \\
Edit-As-Act & Goal-regressive editing loop & Initial scene + instruction via action interface \\
SceneReVis & Multi-turn revision loop & Batch pipeline with scene snapshots \\
\bottomrule
\end{tabularx}
\end{table}

\subsection{Human evaluation interface}
\label{app:human_eval}

This section shows the questionnaire interfaces used for our human evaluation setup. The first interface presents the source scene and the edited result side by side, then asks raters to score the edited output along instruction satisfaction, preservation, physical plausibility, and overall preference. The second interface presents four candidate outputs for the same editing case and asks raters to make an overall forced-choice decision.

\begin{figure*}[t]
\centering
\begin{minipage}[t]{0.48\textwidth}
  \centering
  {\small\textbf{(a) Result-level rating form.}}\par\vspace{0.25em}
  \includegraphics[width=\linewidth,height=0.88\textheight,keepaspectratio]{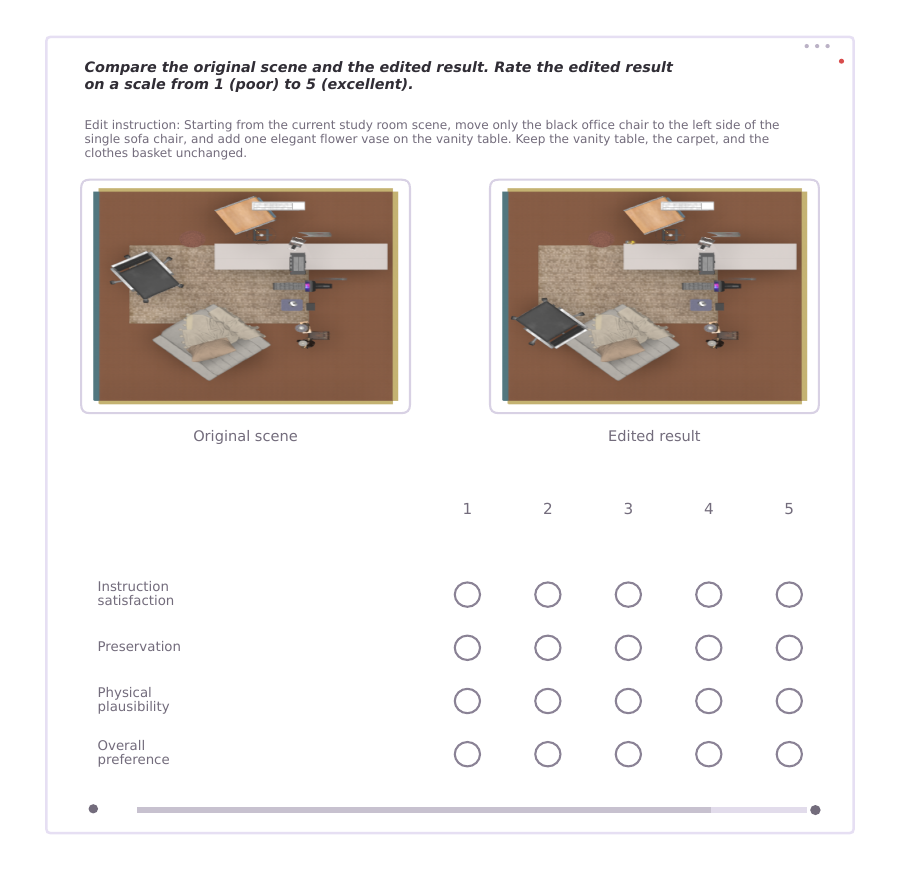}
\end{minipage}\hfill
\begin{minipage}[t]{0.48\textwidth}
  \centering
  {\small\textbf{(b) Four-way forced-choice form.}}\par\vspace{0.25em}
  \includegraphics[width=\linewidth,height=0.88\textheight,keepaspectratio]{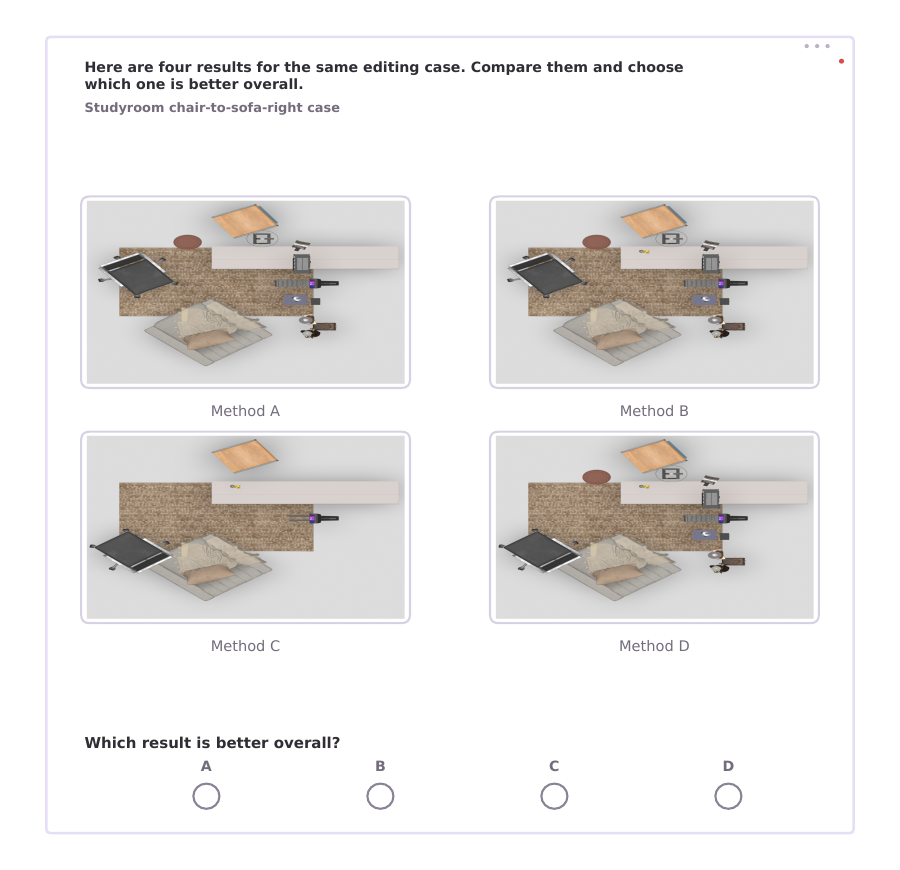}
\end{minipage}
\caption{\textbf{Human evaluation interfaces for preservation-aware editing.} Panel (a) asks annotators to compare the original scene and the edited result, then rate the edited output on instruction satisfaction, preservation, physical plausibility, and overall preference. Panel (b) presents four candidate results for the same editing case and asks annotators to select the overall best result.}
\label{fig:human_eval_interface}
\end{figure*}

\section{Additional Experimental Results}
\label{app:additional_results}

\begin{figure}[t]
\centering
\includegraphics[width=0.72\textwidth]{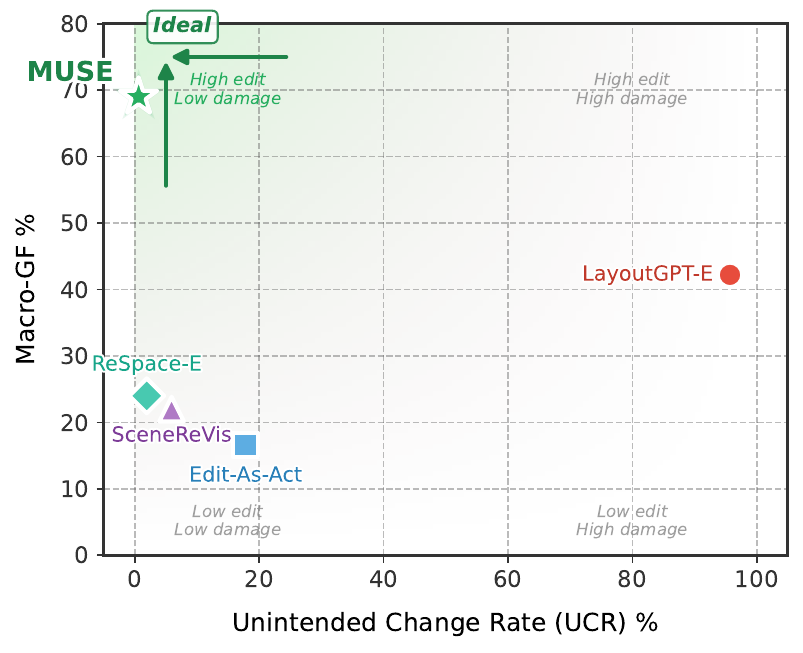}
\caption{\textbf{Completion-locality trade-off.} Higher Macro-GF with lower UCR is preferred. \textsc{MUSE} improves completion while remaining in the low-unintended-change region.}
\label{fig:editing_tradeoff}
\end{figure}

\subsection{Additional qualitative results}
\label{app:qualitative}

\begin{figure}[t]
\centering
\includegraphics[width=0.85\textwidth]{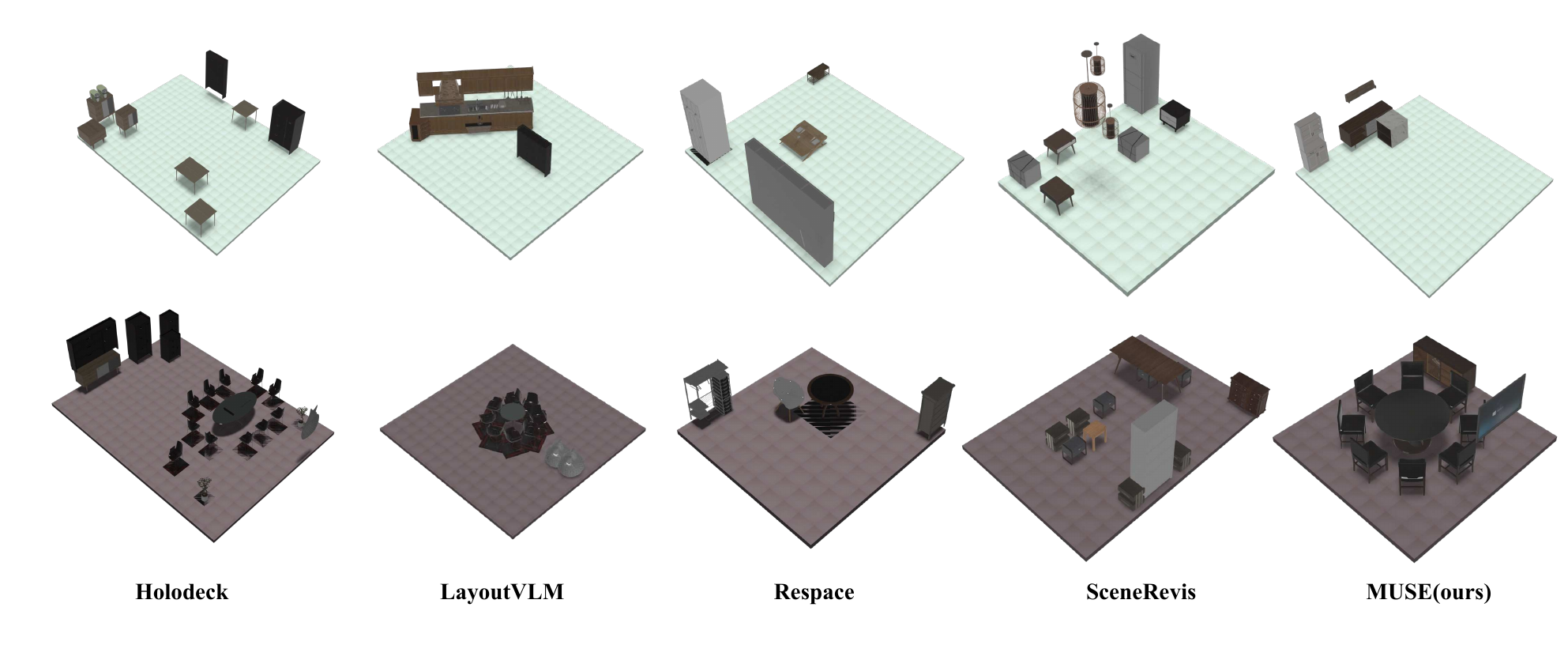}
\vspace{0.5em}
\caption{\textbf{Additional construction comparisons.} \textsc{MUSE} better satisfies compositional requirements while maintaining coherent room layouts.}
\label{fig:construction_qual_appendix}
\end{figure}

\begin{figure}[t]
\centering
\includegraphics[width=0.85\textwidth]{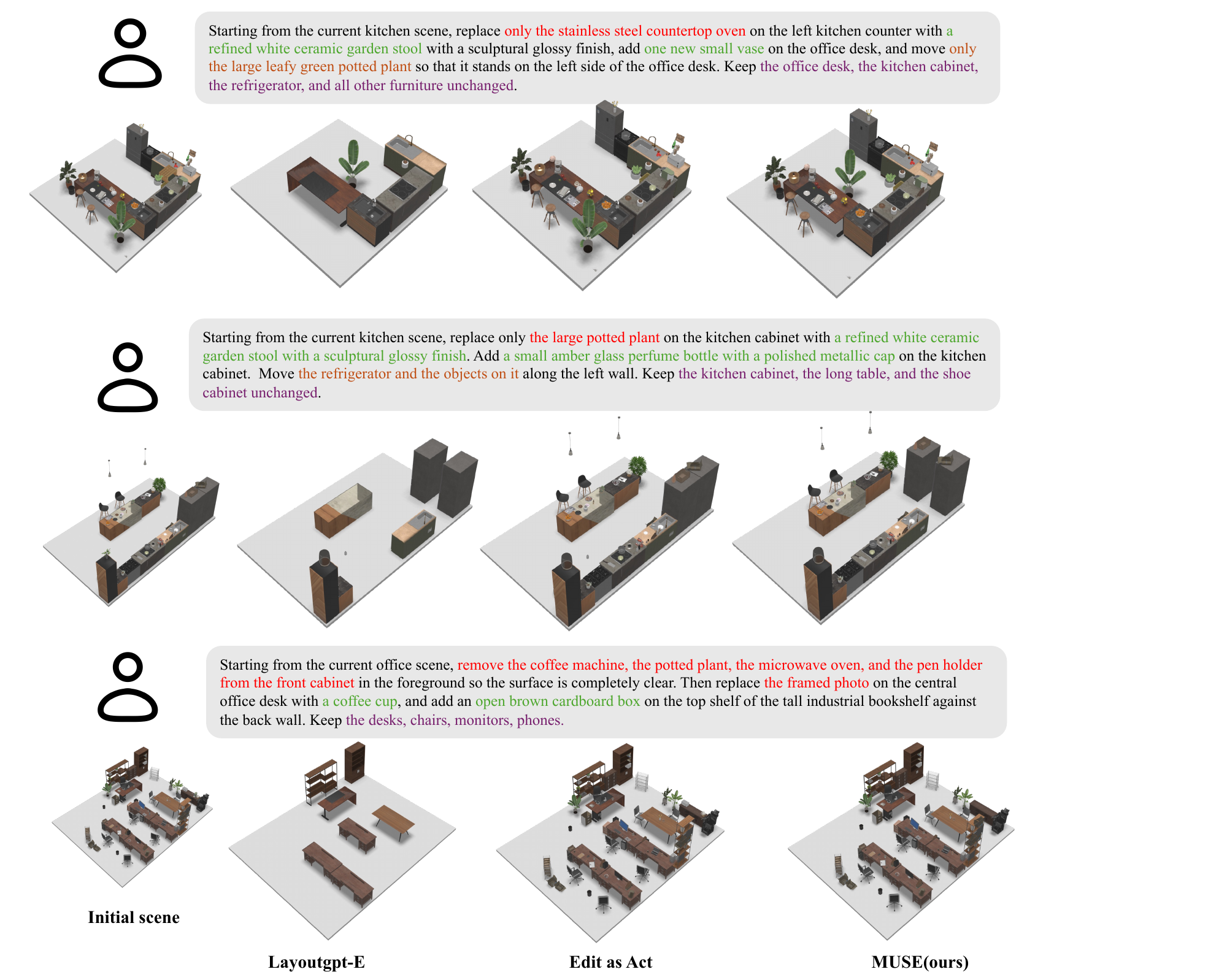}
\vspace{0.5em}
\caption{\textbf{Additional editing comparisons.} \textsc{MUSE} better maintains preservation constraints while completing requested local changes.}
\label{fig:editing_qual_appendix}
\end{figure}

Figure~\ref{fig:construction_qual_appendix} and Figure~\ref{fig:editing_qual_appendix} supplement the main-paper qualitative comparison with additional representative cases from AuthorBench, demonstrating \textsc{MUSE}'s ability to satisfy compositional spatial and support constraints while maintaining local editing precision and preservation.

\subsection{Continuous editing stress test}
\label{app:continuous_stress}

We evaluate \textsc{MUSE} in a continuous editing stress test containing 50 three-stage editing chains (150 stage-level tasks total). Stage 1 constructs the initial scene; stages 2 and 3 apply local edits to the previous output. \textsc{MUSE} completes 139/150 stages (92.7\%); restricting to the 100 continuous editing stages, it completes 89 edits (89.0\%). At the chain level, 39/50 editing chains complete both edit stages successfully. Add and scale operations are robust, while rotate, replace, and move remain harder due to precise target grounding and fine-grained geometric control requirements.

\clearpage

\subsection{Complexity robustness}
\label{app:complexity_robustness}

Construction and editing cases are organized into four tiers: Easy, Medium, Hard, and Complex. We report GF for each tier, then compute retention:
\[
    \mathrm{Ret.} = \frac{\mathrm{GF}_{Complex}}{\mathrm{GF}_{Easy}}.
\]
Higher retention indicates better robustness as compositional complexity increases. A retention above 100\% indicates that the method performs better on Complex cases than on Easy cases.

\begin{figure*}[t]
  \centering
  \includegraphics[width=0.94\textwidth]{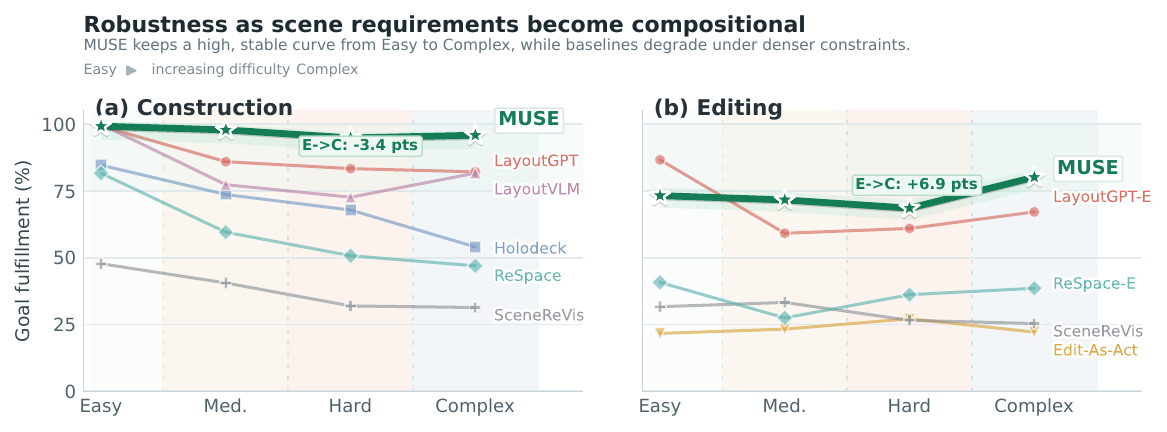}
  \caption{\textbf{Complexity robustness across difficulty tiers.} We plot goal fulfillment for each method from Easy to Complex cases in construction and preservation-aware editing. Higher and flatter curves indicate better robustness as the number and interaction of requirements increase.}
  \label{fig:complexity_robustness_line}
\end{figure*}

\subsection{Downstream navigation curriculum}
\label{app:navigation_curriculum}

The downstream curriculum experiment evaluates whether scene sequences remain spatially coherent as task difficulty increases. Each of 100 cases contains three stages in the same room. The incremental variant edits the previous stage with \textsc{MUSE}; the regeneration variant creates each stage independently. Metrics are computed from exported layouts via a 0.1m occupancy grid with 0.05m obstacle buffer. Table~\ref{tab:navigation_full_stage_metrics} shows stage-wise metrics. Both variants produce navigable final scenes, but \textsc{MUSE} yields stronger spatial continuity with higher difficulty monotonicity (0.850 vs.\ 0.394) and free-space preservation (0.978 vs.\ 0.649).

\begin{table}[t]
\centering
\caption{\textbf{Stage-wise navigation curriculum metrics.} Incremental \textsc{MUSE} and stage-wise regeneration both produce navigable scenes, but incremental editing better preserves free space and difficulty progression.}
\label{tab:navigation_full_stage_metrics}
\footnotesize
\setlength{\tabcolsep}{4pt}
\renewcommand{\arraystretch}{1.05}
\begin{tabular}{@{}llccccc@{}}
\toprule
\textbf{Method} & \textbf{Stage} & \textbf{Walk.} & \textbf{Reach.} & \textbf{Path m} & \textbf{Col.} & \textbf{OOB} \\
\midrule
Incremental \textsc{MUSE} & 1 & 0.906 & 1.000 & 1.235 & 0.000 & 0.000 \\
Incremental \textsc{MUSE} & 2 & 0.886 & 1.000 & 1.554 & 0.000 & 0.000 \\
Incremental \textsc{MUSE} & 3 & 0.868 & 0.992 & 1.736 & 0.000 & 0.000 \\
\midrule
Stage-wise regeneration & 1 & 0.912 & 1.000 & 1.233 & 0.000 & 0.000 \\
Stage-wise regeneration & 2 & 0.894 & 1.000 & 1.794 & 0.000 & 0.000 \\
Stage-wise regeneration & 3 & 0.878 & 0.999 & 2.183 & 0.000 & 0.000 \\
\bottomrule
\end{tabular}
\end{table}

\subsection{Runtime, API cost, and export availability}
\label{app:compute}

All reported results are inference-only: we do not train or fine-tune on AuthorBench. Experiments were run on internal GPU servers for model execution, together with CPU-side scene conversion, rendering, and rule-based checking. The main cost comes from repeated planner / verifier calls, scene export, and final evaluation across the 145 construction cases, the fixed 240-case editing split, and the ablation settings.

Table~\ref{tab:editing_efficiency_profile} provides a compact reference efficiency profile for the editing setting. We report a selected subset of methods with directly comparable single-turn or iterative call structure and sufficiently stable per-case logging, so the table stays readable and focuses on the main runtime trend. We group cases into simple edits (the easy tier) and hard edits (the union of the hard and complex tiers).

\begin{table}[t]
\centering
\small
\setlength{\tabcolsep}{5pt}
\begin{tabular}{lcccc}
\toprule
Method & Avg LLM calls & Avg turns & Avg time / case & Est. API cost / case \\
\midrule
\multicolumn{5}{c}{Simple cases (easy)} \\
Edit-As-Act & $\sim$1 & 1 & $\sim$45s & $\sim$\$0.04 \\
LayoutGPT-E & $\sim$1 & 1 & $\sim$30s & $\sim$\$0.05 \\
MUSE & 3.87 & 0.82 & 260s & \$0.147 \\
\midrule
\multicolumn{5}{c}{Hard cases (hard + complex)} \\
Edit-As-Act & $\sim$1 & 1 & $\sim$90s & $\sim$\$0.08 \\
LayoutGPT-E & $\sim$1 & 1 & $\sim$60s & $\sim$\$0.09 \\
MUSE & 9.10 & 3.27 & 574s & \$0.426 \\
\bottomrule
\end{tabular}
\caption{\textbf{Reference efficiency profile for selected editing methods.} MUSE values are measured from editing-run logs on the 240-case AuthorBench split. Edit-As-Act and LayoutGPT-E are reference estimates under stateless single-turn adaptation and should be interpreted as approximate efficiency profiles rather than exact billing logs. We omit ReSpace-E and SceneReVis from this compact table because their released pipelines expose less directly comparable runtime logging and status granularity; output availability for all editing methods is reported separately in Table~\ref{tab:final_scene_export_rate}. Estimated API cost reflects text-only prompt/response usage and may not fully account for image-token charges.}
\label{tab:editing_efficiency_profile}
\end{table}

Wall-clock time varies across baselines because they expose different released pipelines and different numbers of model calls. We therefore report unified evaluation outcomes rather than a single cross-method runtime claim. Table~\ref{tab:editing_efficiency_profile} should be read as a cost-profile comparison: single-turn baselines stay cheap because they typically make one stateless edit call, whereas MUSE spends more calls and more repair turns to support iterative grounding, verification, and correction. The total project compute exceeded the final reported runs because prompt iteration, dataset curation, debugging, and pilot evaluations were performed before freezing the benchmark protocol and final tables.

Because released baselines expose heterogeneous runtime status labels, we also report a simple output-availability metric: whether a case emits an evaluable \texttt{final\_scene.json} in the common AuthorBench format. This \emph{final-scene export rate} measures generation availability rather than requirement satisfaction, and therefore complements rather than replaces the All-Goal success values in the main paper.

\begin{table}[t]
\centering
\small
\setlength{\tabcolsep}{5pt}
\begin{tabular}{llcc}
\toprule
Task & Method & Final scenes & Export rate \\
\midrule
Construction & LayoutGPT & 145/145 & 100.0\% \\
Construction & Holodeck & 145/145 & 100.0\% \\
Construction & LayoutVLM & 141/145 & 97.2\% \\
Construction & ReSpace & 145/145 & 100.0\% \\
Construction & SceneReVis & 99/145 & 68.3\% \\
Construction & MUSE & 145/145 & 100.0\% \\
\midrule
Editing & LayoutGPT-E & 240/240 & 100.0\% \\
Editing & Edit-As-Act & 212/240 & 88.3\% \\
Editing & ReSpace-E & 240/240 & 100.0\% \\
Editing & SceneReVis & 229/240 & 95.4\% \\
Editing & MUSE & 240/240 & 100.0\% \\
\bottomrule
\end{tabular}
\caption{\textbf{Final-scene export rate on AuthorBench.} A case counts as export-successful if the released pipeline emits an evaluable \texttt{final\_scene.json} in the common AuthorBench format. This is an output-availability metric, not a task-success metric: a run may export a final scene even when it stops early or fails to satisfy all requirements.}
\label{tab:final_scene_export_rate}
\end{table}

\section{Implementation, Assets, and Prompts}
\label{app:implementation_details}

\subsection{Requirement enrichment and skill memory}
\label{app:enrichment}

During requirement compilation, the Architect automatically detects incomplete dependency chains. A requirement with a placement constraint (e.g., ``place a lamp on the nightstand'') implicitly depends on the referenced object's existence. If no requirement declares that object, the system injects an implicit entity requirement with higher priority, then updates the placement requirement's \texttt{depends\_on\_requirement\_ids}. This ensures the focus scheduler processes entity requirements before dependent placements.

Skill Memory $\mathcal{K}$ stores compact motif cards: \texttt{motif\_type}, trigger phrases, expected anchors, and guidance on decomposition, role bindings, and negative hints. Card retrieval scores candidates by trigger-phrase overlap with $I_t$ and anchor-pattern overlap with scene objects, ranks by score, applies mutual-exclusion filtering, and injects the selected cards' guidance into the Architect's prompt. Outside benchmark evaluation, successful authoring episodes can be mined offline to expand the library. In all reported experiments, the library is frozen at test time.

\subsection{Asset library and retrieval}
\label{app:asset_library}

MUSE decouples object identity from spatial placement. The Architect and Sculptor emit object descriptions and size hints, while spatial grounding is handled by placement tools and verification. Our unified asset registry merges the 3D-FUTURE pool with optional sources into a searchable catalog (66,655 records total). Each record stores asset key, source label, bounding-box size, caption, tags, optional category, and embedding indices.

For a query $q$, candidate asset $a$, and target size $\mathbf{s}^{*}$, the retriever ranks assets using:
\[
\mathrm{score}(a \mid q, \mathbf{s}^{*}) =
\lambda_{\text{sbert}} \cos(\mathbf{e}^{\text{text}}_q, \mathbf{e}^{\text{text}}_a)
~+~
\lambda_{\text{clip}} \max_{v} \cos(\mathbf{e}^{\text{clip}}_q, \mathbf{e}^{\text{clip}}_{a,v})
~-~
\lambda_{\text{size}} \cdot \frac{1}{3} \sum_{d \in \{x,y,z\}}
\frac{|s_{a,d} - s^{*}_{d}|}{\max(s^{*}_{d}, \epsilon)}.
\]
Spatial instructions (``on the nightstand'', ``against the east wall'') are excluded from retrieval text and enforced later by placement tools. MUSE retrieves top-$k=20$ candidates; deterministic evaluation uses greedy top-1 selection. After selection, the scene object stores the chosen asset ID, source, bounding box, render path, and a selection breakdown for debugging.

\subsection{Prompt templates}
\label{app:prompt_templates}

The production prompts enumerate output schemas, allowed predicates, tool signatures, and safety constraints. Table~\ref{tab:prompt_template_summary} summarizes the agent-specific behavioral contracts. The Architect converts user requests into structured \texttt{DesignBrief} JSON, enforcing atomic decomposition, verifier-friendly normalization, and memory-guided parsing. The Sculptor plans tool calls respecting protection constraints and scope locality. The Inspector performs per-requirement judging and emits termination signals. The full runtime message assembles the system prompt with few-shot examples, motif cards, scene summaries, protection sets, and rendered images as context.

\begin{table}[t]
\centering
\caption{\textbf{Prompt-contract coverage in the appendix.} We expose one representative prompt excerpt directly and summarize the remaining agent-specific behavioral contracts here, leaving bulky schemas and serialized runtime state to the released code and artifacts.}
\label{tab:prompt_template_summary}
\footnotesize
\setlength{\tabcolsep}{4pt}
\renewcommand{\arraystretch}{1.08}
\begin{tabularx}{\linewidth}{@{}p{0.18\linewidth}p{0.28\linewidth}Y@{}}
\toprule
\textbf{Agent} & \textbf{Primary output} & \textbf{Prompt details shown here} \\
\midrule
Architect & \texttt{DesignBrief} JSON & Exact output discipline, atomic decomposition rules, verifier normalization policy, memory-guided parsing rules \\
Sculptor & \texttt{<tool\_calls>} JSON block & Protection constraints, high-level tool preference, local repair policy, anti-repeat and scope-locality rules \\
Inspector & feedback JSON & Per-requirement judging contract, issue emission rules, termination criteria, model-judged verification policy \\
\bottomrule
\end{tabularx}
\end{table}

\subsection{Interactive frontend}
\label{app:frontend}

Beyond offline benchmark export, MUSE is exposed through an interactive frontend for replaying AuthorBench cases and inspecting intermediate system state. The interface provides complementary views: the main render workspace for loaded cases, the linked layout inspector with object-level geometry, and the memory/verification panel that surfaces requirement progress, protection sets, and working-memory state.

\clearpage
\section{Limitations and Broader Impacts}
\label{app:limitations_impacts}

\subsection{Failure mode analysis}
\label{app:failure_modes}

We summarize the main failure modes observed in incomplete or partially completed runs. First, fine-grained local transforms remain the most fragile operation family. In these cases, the system usually preserves the existing scene and often grounds the target object, but the final movement, scale, or orientation constraint can remain unsatisfied after repeated repair attempts. Rotation is the clearest example: several runs pass coarse target checks but fail the verifier because the object does not precisely face the requested reference.

Second, support-surface and wall-grounding constraints are difficult when they interact with room layout constraints. Construction cases that combine object inventory, support placement, wall anchoring, and zone composition can fail even when the requested object categories are mostly present, because satisfying all spatial predicates jointly requires precise placement and later feasibility repair. Third, compound edits are more error-prone than atomic edits. A replacement or insertion may succeed at the object level, while a dependent relation such as ``near'', ``left of'', ``on top of'', or ``facing'' remains unresolved. Finally, a small number of cases fail during residual physical cleanup, showing that semantic satisfaction and geometric feasibility are not always aligned in dense scenes. These patterns indicate that persistent scene state improves local continuity, but high-precision geometric control and multi-constraint reconciliation remain important directions for future work.

\subsection{Limitations}
\label{app:limitations}

MUSE has several limitations. First, the main benchmark focuses on indoor scenes and single-instruction edits; the supplementary continuous-editing stress test only covers controlled three-stage chains, so the current results do not establish performance on outdoor assets, open-ended multi-turn authoring sessions, or substantially different scene ontologies. Second, MUSE relies on repeated LLM planning and verification calls; when the model misunderstands a requirement or proposes a poor intermediate action, the loop can incur extra latency or terminate with only partial progress. Third, the pipeline assumes access to a structured scene representation and a sufficiently rich asset pool, which may not hold in settings with noisy geometry, missing metadata, or sparse object libraries. These limitations motivate broader-domain benchmarks and more efficient verification and memory mechanisms.

\subsection{Broader impacts}
\label{app:broader_impacts}

Controllable 3D scene authoring can benefit embodied-AI simulation, interactive environment design, and assistive layout tools by making requirements explicit and auditable. At the same time, systems of this kind can produce unsafe, unrealistic, or misleading layouts if they are used without human review or if downstream users over-trust partially satisfied outputs.

Our mitigation emphasis is on verifiability rather than autonomy. MUSE exposes requirement-level checks, reports preservation and unintended-change metrics for editing, and is evaluated with external benchmark checks that are independent from the system's internal verifier. Even so, human oversight remains necessary when generated scenes are used for training, design, or any downstream decision-making setting.


{\small
\begin{thebibliography}{99}

\bibitem{tang2024scenegeneration}
Tang, J., Nie{\ss}ner, M., \& Dai, A. (2024).
DiffuScene: Denoising diffusion models for generative indoor scene synthesis.
In \textit{Proc. CVPR}.

\bibitem{paschalidou2021atiss}
Paschalidou, D., Kar, A., Shugrina, M., Kreis, K., Geiger, A., \& Fidler, S. (2021).
ATISS: Autoregressive transformers for indoor scene synthesis.
In \textit{Proc. NeurIPS}.

\bibitem{yang2024holodeck}
Yang, Y., Fan, R., Xu, J., Yu, M., Akkaynak, D., Gao, R., \& Ilie, A. (2024).
Holodeck: Language guided generation of 3D embodied AI environments.
In \textit{Proc. CVPR}.

\bibitem{fu20213dfront}
Fu, H., Cai, B., Gao, L., Zhang, L. X., Wang, J., Li, C., Zeng, Q., Sun, C., Jia, R., Zhao, B., \& Zhang, L. (2021).
3D-FRONT: 3D furnished rooms with layouTs and semantics.
In \textit{Proc. ICCV}.

\bibitem{wang2021sceneformer}
Wang, X., Yeshwanth, C., \& Nie{\ss}ner, M. (2021).
SceneFormer: Indoor scene generation with transformers.
In \textit{Proc. 3DV}.

\bibitem{dhamo2021graphto3d}
Dhamo, H., Manhardt, F., Navab, N., \& Tombari, F. (2021).
Graph-to-3D: End-to-End Generation and Manipulation of 3D Scenes Using Scene Graphs.
In \textit{Proc. ICCV}.

\bibitem{feng2024layoutgpt}
Feng, W., Zhu, W., Fu, T.-J., Jampani, V., Akula, A. R., He, X., Basu, S., Wang, X. E., \& Wang, W. Y. (2023).
LayoutGPT: Compositional Visual Planning and Generation with Large Language Models.
In \textit{Proc. NeurIPS}.

\bibitem{sun2025layoutvlm}
Sun, F. Y., Liu, W., Gu, S., Lim, D., Bhat, G., Tombari, F., Li, M., Haber, N., \& Wu, J. (2025).
LayoutVLM: Differentiable Optimization of 3D Layout via Vision-Language Models.
In \textit{Proc. CVPR}.

\bibitem{chaplot2020objgoal}
Chaplot, D. S., Gandhi, D., Gupta, S., Gupta, A., \& Salakhutdinov, R. (2020).
Learning to explore using active neural SLAM.
In \textit{Proc. ICLR}.

\bibitem{gu2023conceptgraphs}
Gu, Q., Kuwajerwala, A., Jatavallabhula, K. M., Sen, B., Agarwal, A., Rivera, C., Paul, W., Chellappa, R., Gan, C., de Melo, C. M., Tenenbaum, J. B., Torralba, A., Shkurti, F., \& Paull, L. (2024).
ConceptGraphs: Open-Vocabulary 3D Scene Graphs for Perception and Planning.
In \textit{Proc. ICRA}.

\bibitem{wang2023voyager}
Wang, G., Xie, Y., Jiang, Y., Mandlekar, A., Xiao, C., Zhu, Y., Fan, L., \& Anandkumar, A. (2023).
Voyager: An open-ended embodied agent with large language models.
\textit{arXiv preprint}.

\bibitem{park2023generativeagents}
Park, J. S., O'Brien, J. C., Cai, C. J., Morris, M. R., Liang, P., \& Bernstein, M. S. (2023).
Generative agents: Interactive simulacra of human behavior.
In \textit{Proc. UIST}.

\bibitem{sumers2023coala}
Sumers, T. R., Yao, S., Narasimhan, K., \& Griffiths, T. L. (2023).
Cognitive architectures for language agents.
\textit{arXiv preprint arXiv:2309.02427}.

\bibitem{lin2024instructscene}
Lin, C., Xu, J., Zhu, J., Liang, Y., Zhang, L., \& Huang, S. (2024).
InstructScene: Instruction-driven 3D indoor scene synthesis with semantic graph prior.
In \textit{Proc. ICLR}.

\bibitem{yang2025sceneweaver}
Yang, Y., Jia, B., Zhang, S., \& Huang, S. (2025).
SceneWeaver: All-in-One 3D Scene Synthesis with an Extensible and Self-Reflective Agent.
In \textit{Proc. NeurIPS}.

\bibitem{zhu2025imaginarium}
Zhu, X., Huang, X., Xie, Q., Deng, Z., Yu, J., Guan, Y., Liu, Z., Zhu, L., Zhao, Q., Liu, L., \& Zeng, L. (2025).
Imaginarium: Vision-Guided High-Quality 3D Scene Layout Generation.
\textit{ACM Transactions on Graphics}, 44(6).

\bibitem{wu2026scenesmith}
Pfaff, N., Cohn, T., Zakharov, S., Cory, R., \& Tedrake, R. (2026).
SceneSmith: Agentic Generation of Simulation-Ready Indoor Scenes.
\textit{arXiv preprint arXiv:2602.09153}.

\bibitem{huang2026editasact}
Noh, S., Seo, S., Park, G.-M., \& Kang, H. (2026).
Edit-As-Act: Goal-Regressive Planning for Open-Vocabulary 3D Indoor Scene Editing.
In \textit{Proc. CVPR}.

\bibitem{zhang2026sage}
Xia, H., Li, X., Li, Z., Ma, Q., Xu, J., Liu, M.-Y., Cui, Y., Lin, T.-Y., Ma, W.-C., Wang, S., Song, S., \& Wei, F. (2026).
SAGE: Scalable Agentic 3D Scene Generation for Embodied AI.
\textit{arXiv preprint arXiv:2602.10116}.

\bibitem{pun2026hsm}
Pun, H. I. D., Tam, H. I., Wang, A. T., Huo, X., Chang, A. X., \& Savva, M. (2026).
HSM: Hierarchical Scene Motifs for Multi-Scale Indoor Scene Generation.
In \textit{Proc. 3DV}.



\bibitem{yang2024physcene}
Yang, Y., Jia, B., Zhi, P., \& Huang, S. (2024).
PhyScene: Physically Interactable 3D Scene Synthesis for Embodied AI.
In \textit{Proc. CVPR}.

\bibitem{fu2024anyhome}
Fu, R., Wen, Z., Liu, Z., \& Sridhar, S. (2024).
AnyHome: Open-vocabulary generation of structured and textured 3D homes.
In \textit{Proc. ECCV}.

\bibitem{gu2025artiscene}
Gu, Z., Cui, Y., Li, Z., Wei, F., Ge, Y., Gu, J., Liu, M.-Y., Davis, A., \& Ding, Y. (2025).
ArtiScene: Language-driven artistic 3D scene generation through image intermediary.
In \textit{Proc. CVPR}.

\bibitem{shinn2023reflexion}
Shinn, N., Cassano, F., Gopinath, A., Narasimhan, K., \& Yao, S. (2023).
Reflexion: Language agents with verbal reinforcement learning.
In \textit{Proc. NeurIPS}.

\bibitem{zhao2026scenerevis}
Zhao, Y., Sun, S., Zhang, M., Shi, Y., Yang, X., \& Bian, J. (2026).
SceneReVis: A self-reflective vision-grounded framework for 3D indoor scene synthesis via multi-turn RL.
\textit{arXiv preprint arXiv:2602.09432}.
\bibitem{yao2023react}
Yao, S., Zhao, J., Yu, D., Du, N., Shafran, I., Narasimhan, K., \& Cao, Y. (2023).
ReAct: Synergizing reasoning and acting in language models.
In \textit{Proc. ICLR}.

\bibitem{huang2022innermonologue}
Huang, W., Xia, F., Xiao, T., Chan, H., Liang, J., Florence, P., Zeng, A., Tompson, J., Mordatch, I., Chebotar, Y., Sermanet, P., Brown, T., Jackson, T., Luu, L., Levine, S., Hausman, K., \& Ichter, B. (2022).
Inner monologue: Embodied reasoning through planning with language models.
In \textit{Proc. CoRL}.

\bibitem{yang2024llplace}
Yang, Y., Lu, J., Zhao, Z., Luo, Z., Yu, J. J. Q., Sanchez, V., \& Zheng, F. (2024).
LLplace: The 3D Indoor Scene Layout Generation and Editing via Large Language Model.
\textit{arXiv preprint arXiv:2406.03866}.

\bibitem{wang2025chat2layout}
Wang, C., Zhong, H., Chai, M., He, M., Chen, D., \& Liao, J. (2024).
Chat2Layout: Interactive 3D Furniture Layout with a Multimodal LLM.
\textit{arXiv preprint arXiv:2407.21333}.

\bibitem{bucher2025respace}
Bucher, M. J. J., \& Armeni, I. (2025).
ReSpace: Text-Driven 3D Scene Synthesis and Editing with Preference Alignment.
\textit{arXiv preprint arXiv:2506.02459}.

\bibitem{cheng2026layoutr1}
Zhen, H., Li, X., Zhao, Y., Zhang, H., Liu, S., Mo, K., Gan, C., \& Radhakrishnan, S. (2026).
3D-Layout-R1: Structured Reasoning for Language-Instructed Spatial Editing.
\textit{arXiv preprint arXiv:2603.22279}.

\end{thebibliography}
}
\end{document}